\documentclass{article}

\PassOptionsToPackage{numbers}{natbib}

    \usepackage[preprint]{neurips_2020}

\usepackage[utf8]{inputenc} 
\usepackage[T1]{fontenc}    
\usepackage{hyperref}       
\usepackage{url}            
\usepackage{booktabs}       
\usepackage{amsfonts}       
\usepackage{nicefrac}       
\usepackage{microtype}      
\usepackage{amsmath}
\usepackage{amssymb}
\usepackage[tbtags]{mathtools}

\usepackage{caption}
\usepackage{subcaption}
\usepackage{makecell}

\usepackage{hyperref}
\usepackage{url}
\usepackage[ruled,vlined,noend,linesnumbered]{algorithm2e}
\usepackage{multirow}
\usepackage{graphicx}

\usepackage{xcolor}
\DeclareMathOperator*{\argmax}{argmax} 

\title{Accurate and Fast Federated Learning \\ via Combinatorial Multi-Armed Bandits}

\author{%
  Taehyeon Kim\thanks{Equally Contributed.} \\
  KAIST\\
  \texttt{potter32@kaist.ac.kr} \\
  \And
  Sangmin Bae$^*$ \\
  KAIST \\
  \texttt{dkswodus49@kaist.ac.kr} \\
  \AND
  Jin-woo Lee \\
  KAIST \\
  \texttt{jinwoo.lee@kaist.ac.kr} \\
  \And
  Seyoung Yun \\
  KAIST \\
  \texttt{yunseyoung@gmail.com} \\
}

\usepackage{CJKutf8}
\begin{document}
\begin{CJK}{UTF8}{mj} 
\maketitle
\vspace{-10pt}
\begin{abstract}
\vspace{-10pt}
Federated learning has emerged as an innovative paradigm of collaborative machine learning. Unlike conventional machine learning, a global model is collaboratively learned while data remains distributed over a tremendous number of client devices, thus not compromising user privacy. However, several challenges still remain despite its glowing popularity; above all, the global aggregation in federated learning involves the challenge of biased model averaging and lack of prior knowledge in client sampling, which, in turn, leads to high generalization error and slow convergence rate, respectively. In this work, we propose a novel algorithm called \textbf{\textit{FedCM}} that addresses the two challenges by utilizing prior knowledge with multi-armed bandit based client sampling and filtering biased models with combinatorial model averaging. Based on extensive evaluations using various algorithms and representative heterogeneous datasets, we showed that FedCM significantly outperformed the state-of-the-art algorithms by up to $\mathbf{37.25\%}$ and $\mathbf{4.17}$ times, respectively, in terms of generalization accuracy and convergence rate.
\end{abstract}

\section{Introduction}

Federated learning\,(FL)\,\citep{fedavg,konevcny2016federated} enables mobile devices to collaboratively learn a shared model while keeping all training data on the devices, thus avoiding transferring data to the cloud or central server. One of the main reasons for this recent
boom in FL is that it does not compromise
user privacy.
In this framework, a local model is updated via its private data on the corresponding local device; all local updates are aggregated to the global model; after which the procedure is repeated until convergence.

In particular, the canonical \emph{global aggregation} involves sampling clients as well as averaging the models of sampled clients\,\citep{fedavg}. Even though client sampling and model averaging schemes are frequently proposed\,\citep{fedprox,li2019convergence,nishio2019client}, less has been addressed the inherent dynamics of how they influence the global aggregation. To this end, we identified two challenges that arose when the conventional algorithms were used as follows:

\begin{itemize}
\item
    \textbf{Biased Model Averaging}: For the conventional model averaging schemes\,\citep{fedavg,fedprox,li2019convergence}, we identified that the generalization error in non-IID (i.e., independent and identically distributed) setting was not only higher, but also more variant than that of IID setting because the existing schemes did not filter the biased models.
\item
    \textbf{Lack of Prior Knowledge in Client Sampling}: For the conventional client sampling schemes\,\citep{fedavg,fedprox,li2019convergence}, it was observed that the existing schemes led to bad local optima\,\citep{agnosticfl,powerofchoice}, and even the convergence speed was inevitably slow
    since they did not take into account prior knowledge of the client sampling process at all.
\end{itemize}

\begin{table}[t!]
\caption{\textbf{Comparison of algorithms}. + and * sign denotes sampling scheme with and without replacement, respectively. Note that the proposed FedCA and FedCM can be easily extended to any other client sampling and model averaging schemes.}
\centering
\begin{tabular}{ccccc} \toprule
    Algorithm & \makecell{Model \\ Filtering} & \makecell{Prior \\ Knowledge} & \makecell{Sampling \\ Scheme for $S^t$} & \makecell{Averaging \\ Scheme for $w^t$} \\ \midrule
    FedAvg\,\citep{fedavg} & X & X & Uniform* & $\sum_{k \notin S^t} p_{k}w^t + \sum_{k \in S^t} p_{k}w_{k}^{t}$ \\
    FedProx\,\citep{fedprox} & X & X & $p_k$+ & $\frac{1}{|S^t|} \sum_{k \in S^t} w_{k}^{t}$ \\
    FedPdp\footnotemark\,\citep{li2019convergence} & X & X & Uniform* & $\sum_{k \in S^t} p_{k} \frac{|S|}{|S^t|} w_{k}^{t}$ \\
    FedCA\,(Ours) & O & X & Uniform* & $\sum_{k \in S^t_{opt}} p_{k} \frac{|S|}{|S^t_{opt}|} w_{k}^{t}$ \\
    FedCM\,(Ours) & O & O & Bandit* & $\sum_{k \in S^t_{opt}} p_{k} \frac{|S|}{|S^t_{opt}|} w_{k}^{t}$ \\ \bottomrule
\end{tabular}
\vspace{-10pt}
\label{tab:algorithm}
\end{table}
\vspace{-5pt}
\addtocounter{footnote}{-1}
\stepcounter{footnote}\footnotetext{The algorithm proposed by \citet{li2019convergence} is referred to as FedPdp\,(FedAvg with partial device participation).}

To the best of our knowledge, no existing work has addressed both of the above challenges simultaneously, which is shown by \autoref{tab:algorithm} that compares the algorithms from the perspective of model filtering and prior knowledge. To this end, we propose a novel algorithm called \textbf{\textit{FedCM}}\,(\underline{Fed}erated learning with \emph{\underline{C}ombinatorial model averaging} and \emph{\underline{M}ulti-armed bandit\,(MAB) based client sampling}) that resolves both challenges. With \emph{combinatorial model averaging}, we aim to filter biased models in consideration of the model combination that maximizes a validation score, consequently reducing generalization error. In addition, with \emph{MAB based client sampling}, we utilize prior knowledge that models previous client sampling behavior by using a MAB based sampling scheme. The increased information can, in turn, lead to improved convergence performance. Overall, the key contributions are summarized as follows:

\begin{itemize}
\item
    \textbf{Problem Formulation}\,(Section \ref{sec:Problem}): We formulate the problem as a novel system-level framework of \emph{FL with knowledgeable sampling and filtered averaging} that serves as a baseline template for any extension with custom prior knowledge or custom model filter.
\item
    \textbf{Combinatorial Model Averaging}\,(Section \ref{sec:CA}): We design a novel algorithm called \emph{FedCA} to resolve the challenge of biased model averaging. We confirmed that FedCA outperformed the state-of-the-art algorithms by up to $\mathbf{16.75\%}$ in terms of generalization accuracy.
\item
    \textbf{MAB based Client Sampling}\,(Section \ref{sec:MAB}): We finally propose \emph{FedCM} to resolve both challenges. Then, we extensively compared FedCM with various client sampling algorithms for representative heterogeneous datasets. FedCM reached a higher test accuracy by up to $\mathbf{37.25\%}$ as well as a fast convergence rate by up to $\mathbf{4.17\%}$ times.
\end{itemize}

\section{Problem: FL with Knowledgeable Sampling and Filtered Averaging}
\label{sec:Problem}

The objective of \emph{federated learning}\,\citep{fedavg} is to solve the stochastic convex optimization problem:
\begin{gather}
    \min_{w} f(w) \triangleq \sum_{k \in S} p_{k}F_{k}(w)
\end{gather}
where $S$ is the set of total clients, $p_{k}$ is the weight of client $k$, such as $p_{k} \geq 0$, and $\sum_{k}p_{k} =1$. The local objective of client $k$ is to minimize $F_{k}(w) = \mathbb{E}_{x_{k} \sim D_{k}} [\ell_{k}(x_{k},y_{k};w)]$ parameterized by $w$ on the local data $(x_{k},y_{k})$ from local data distribution $D_{k}$.

\emph{FederatedAveraging}\,(FedAvg)\,\citep{fedavg}, the canonical algorithm for FL, involves \emph{local update}, which learns a local model $w^t_{k}$\,(Eq.~\eqref{eq:w_k}) with learning rate $\eta$ and synchronizing $w^t_{k}$ with $w^t$ every $E$ steps,
\begin{equation}
\label{eq:w_k}
\begin{multlined}
    w^{t}_{k} \triangleq \begin{dcases}
        w^{t-1}_{k} - \eta\nabla F_{k}(w^{t-1}_{k}) & \text{if} ~ t \text{ mod } E \ne 0 \\
        w^t & \text{if} ~ t \text{ mod } E = 0
    \end{dcases}
\end{multlined}
\end{equation}

and \emph{global aggregation}, which learns the global model $w^t$\,(Eq.~\eqref{eq:w}) by averaging all $w^t_{k}$ with regard to the client $k \in S^t$ uniformly sampled at random, subject to $|S^t| = SamplingRatio \times |S|$.
\begin{equation}
\label{eq:w}
w^t \triangleq \sum_{k \notin S^t} p_{k}w^t + \sum_{k \in S^t} p_{k}w_{k}^{t}
\end{equation}

\vspace{-10pt} In a similar vein, recent studies\,\citep{fedprox,li2019convergence} proposed different client sampling and model averaging schemes. \autoref{tab:algorithm} summarizes them and shows that prior knowledge and model filter may resolve the challenges. Therefore, by augmenting the client sampling and model averaging scheme with prior knowledge and model filter, respectively, we can derive a generic system-level framework of \emph{FL with knowledgeable sampling and filtered model averaging}, as shown in Algorithm\,\ref{al:Framework}.


\begin{algorithm}[t!]
\SetAlgoLined
\SetKwInOut{Input}{\textsc{Input}}
\SetKwInOut{Output}{\textsc{Output}}
\caption{Generic framework of FL with knowledgeable sampling and filtered averaging}
\label{al:Framework}
\Input{$S, SamplingRatio, PriorKnowledge, ModelFilter, \eta$}
\Output{$w^T$}
Initialize $w^0$ randomly \\
\For{$t \leftarrow 0, \dots , T-1$}{
    $S^t \leftarrow \textsc{SampleClients}(S, SamplingRatio, PriorKnowledge)$ \\
    \For{each client $k \in S^t$ in parallel}{
        $w^t_{k,0} \leftarrow w^t$ \\
        \For{$e \leftarrow 0,\dots, E-1$}{
            $w^{t,e+1}_{k} \leftarrow w^{t,e}_{k} - \eta\nabla F_{k}(w^{t,e}_{k})$ \tcp{Eq.~\eqref{eq:w_k}}
        }
    }
    $w^{t+1} \leftarrow \textsc{AverageModels}(S^t, ModelFilter)$
}

\end{algorithm}

\section{Combinatorial Averaging (CA)}
\label{sec:CA}

In this section, we propose a novel algorithm called \textbf{\textit{FedCA}}(\emph{\underline{Fed}erated learning with \underline{C}ombinatorial \underline{A}veraging}) and systematically evaluate FedCA with various algorithms and representative heterogeneous datasets.

{\bf Proposed Algorithm: FedCA}. To resolve the aforementioned challenge of biased model averaging, we propose FedCA by extending $ModelFilter$ and $AverageModels$ of Algorithm\,\ref{al:Framework} as follows. First, for the $ModelFilter$, we propose \emph{combinatorial model filter} that filters out biased models considering the model combination that maximizes a validation score, which can be expressed as
\begin{gather}
\label{eq:CombOpt}
    S^{t}_{opt} \triangleq \argmax_{\mathcal{S}_{opt}^t \subset S^t} u \left( \mathcal{X}_{val}, \Tilde{g}(x; \mathcal{S}_{opt}^t)\right) ~~ \text{where} ~~ \Tilde{g}(x; S) = \frac{1}{|S|}\sum_{s \in S} g(x_s; w_s)
\end{gather}
where $\mathcal{X}_{val}$ is a validation dataset, $g(x;w)$ is the logit from a network parameterized by $w$, and $u(\mathcal{X}, g)$ is a score function of $u$ on $\mathcal{X}$. In addition, we design two score functions for $u(\mathcal{X}, g)$:
\begin{itemize}
    \item \textbf{Dirac delta function:} It is defined as $\frac{1}{|\mathcal{X}|} \sum_{(x,y) \in \mathcal{X}} \mathbf{1}_{[y = \argmax_c g_c(x; w)]}$ where $\mathbf{1}_{[\cdot]}$ be the indicator function; it is also known as the accuracy.
    \item \textbf{Classification loss:} It is defined as $-\frac{1}{|\mathcal{X}|} \sum_{(x,y) \in \mathcal{X}} \mathbf{log}\, g_y(x; w)$; it is also known as the cross-entropy loss.
\end{itemize}

Next, for the $AverageModels$, we change \hyperref[tab:algorithm]{the state-of-the-art model averaging scheme} of FedPdp by replacing $S^t$ with $S^t_{opt}$ from Eq.~\eqref{eq:CombOpt}, as shown in Eq.~\eqref{eq:w_CA}. In conclusion, FedCA extends $ModelFilter$ to the combinatorial model filter of Eq.~\eqref{eq:CombOpt} and $AverageModels$ to Eq.~\eqref{eq:w_CA}, thus called \emph{combinatorial averaging}. It is apparent that FedCA can be easily extended to any other existing schemes. Please refer to \autoref{app:fedca} for the detailed process illustration of FedCA.
\begin{equation}
\label{eq:w_CA}
w^t \triangleq \sum_{k \in S^t_{opt}} p_{k} \frac{|S|}{|S^t_{opt}|} w_{k}^{t}
\end{equation}

\begin{figure}[t!]
     \centering
     \begin{subfigure}[b]{0.90\textwidth}
         \centering
         \includegraphics[width=\textwidth]{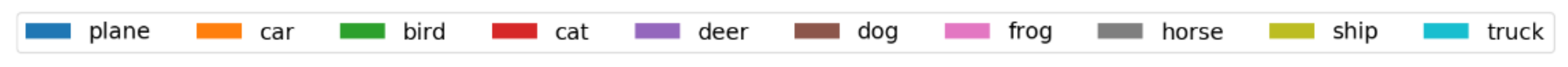}
     \end{subfigure}
     \centering
     \begin{subfigure}[b]{0.32\textwidth}
         \centering
         \includegraphics[width=\textwidth]{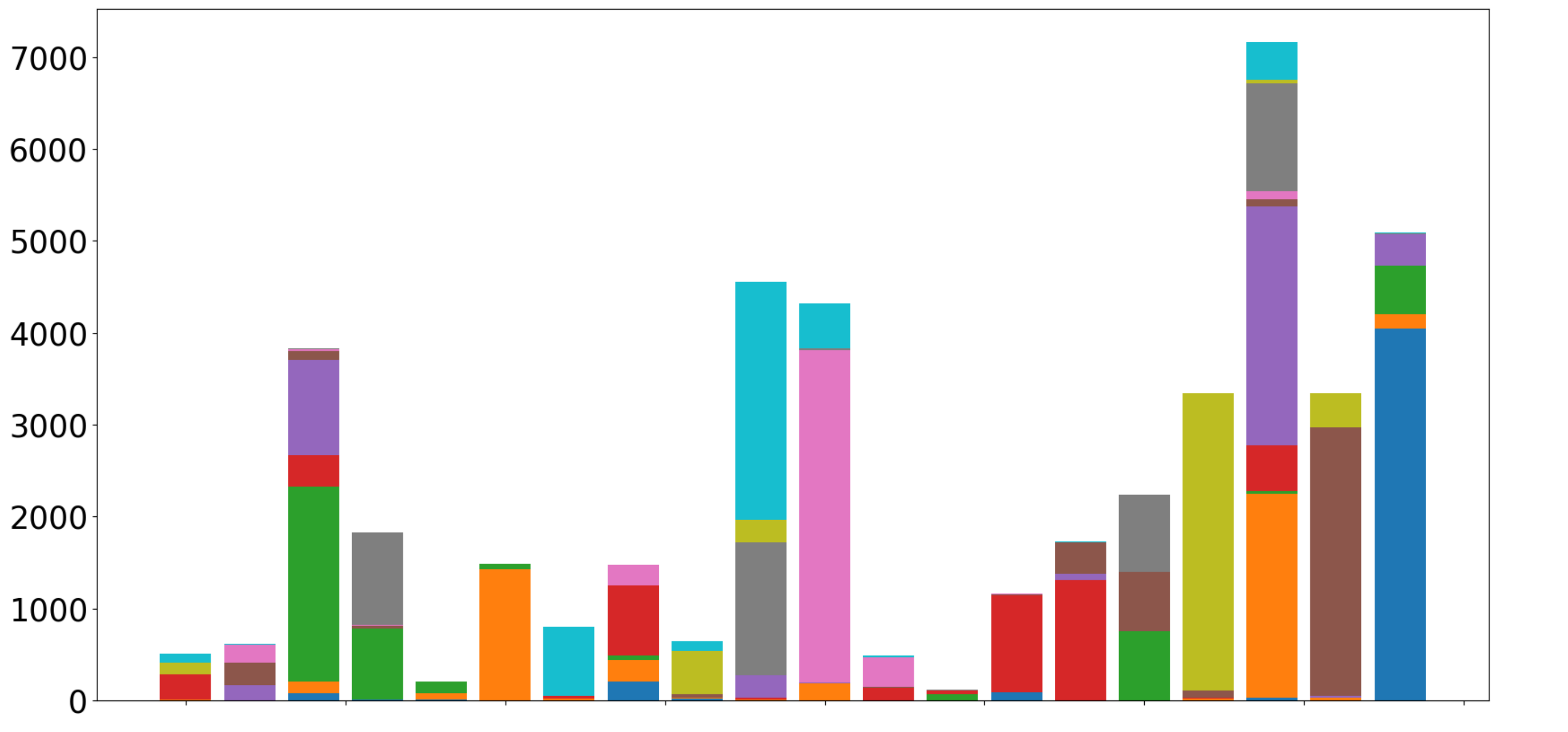}
         \caption{$\alpha=0.1$}
     \end{subfigure}
     \begin{subfigure}[b]{0.32\textwidth}
         \centering
         \includegraphics[width=\textwidth]{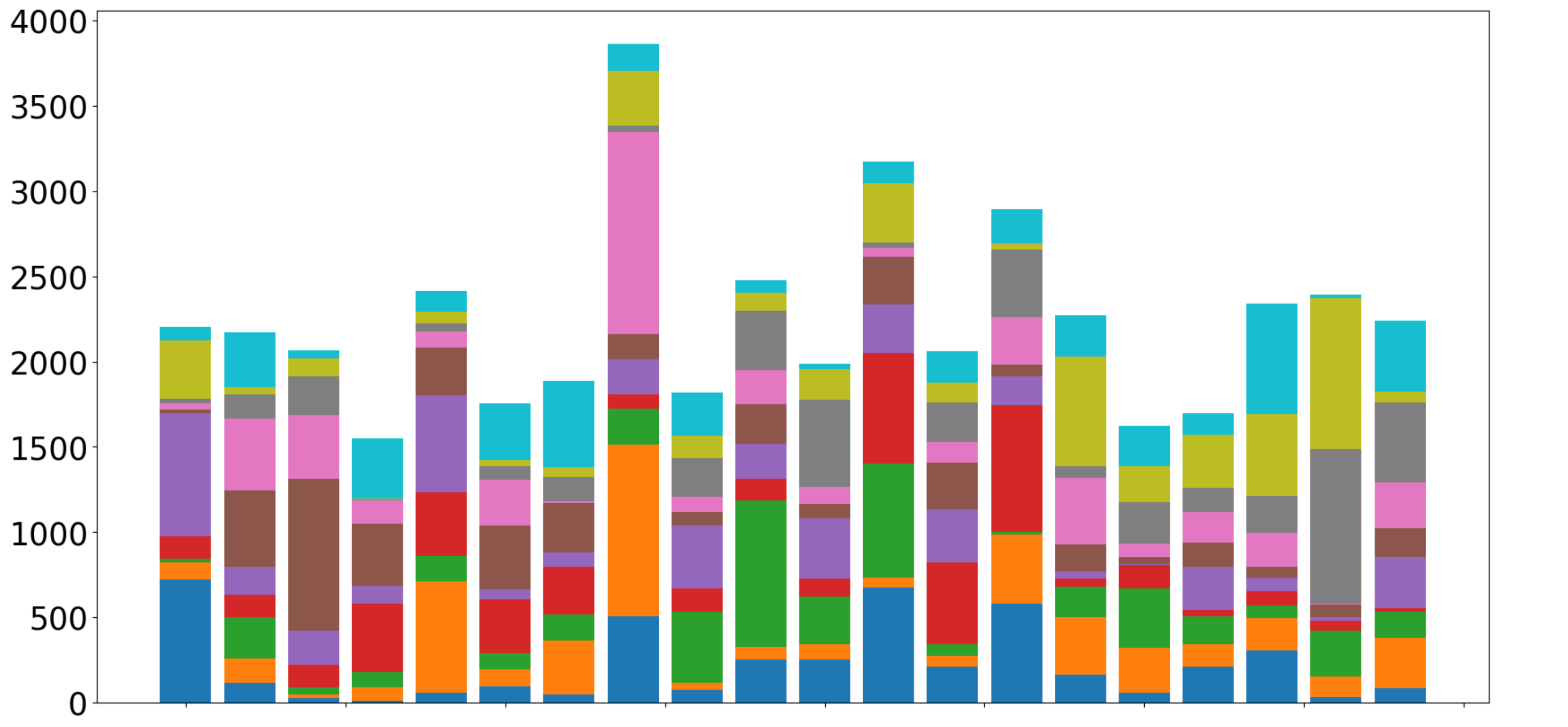}
         \caption{$\alpha=1.0$}
     \end{subfigure}
     \begin{subfigure}[b]{0.32\textwidth}
         \centering
         \includegraphics[width=\textwidth]{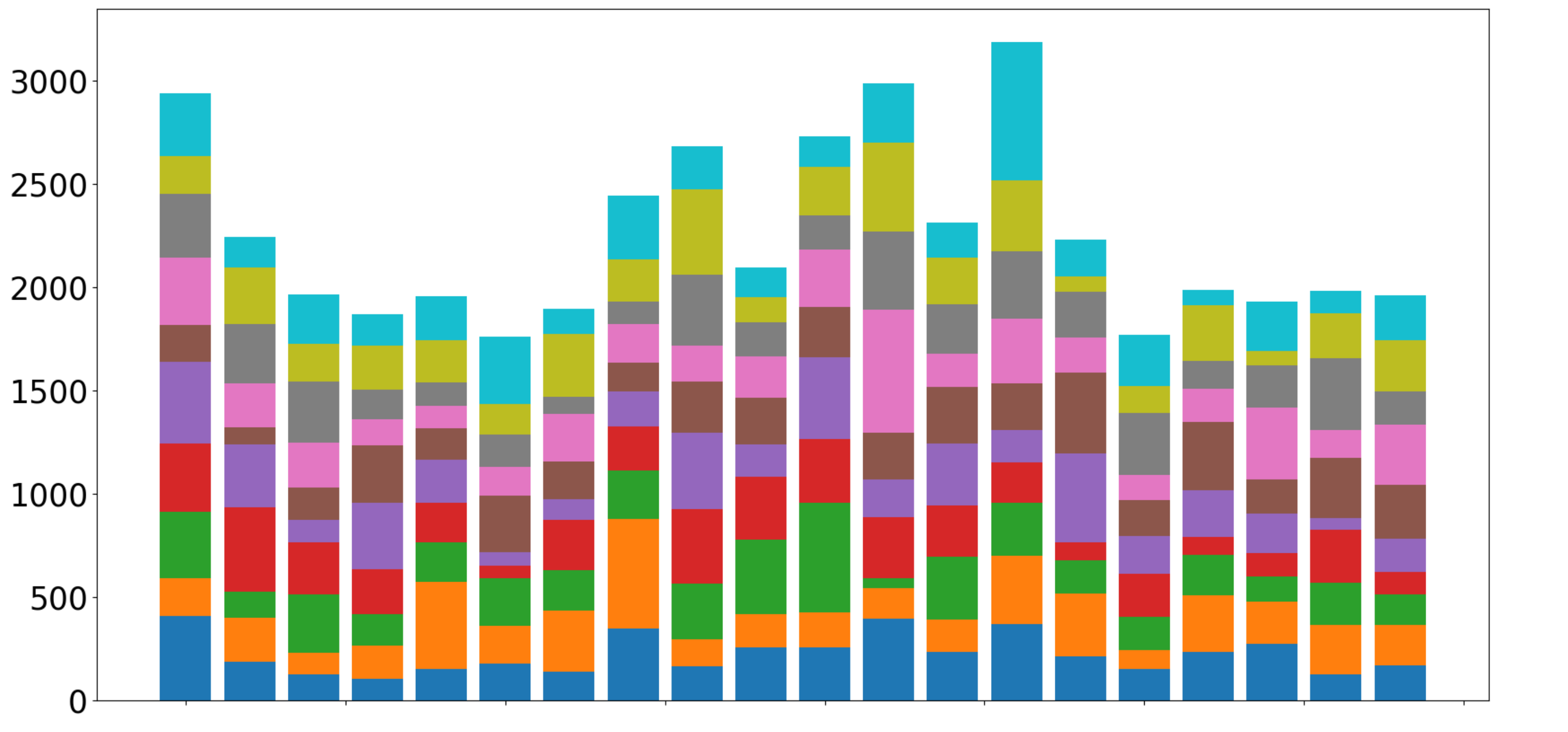}
         \caption{$\alpha=5.0$}
     \end{subfigure}
     \centering
     \begin{subfigure}[b]{0.32\textwidth}
         \centering
         \includegraphics[width=\textwidth]{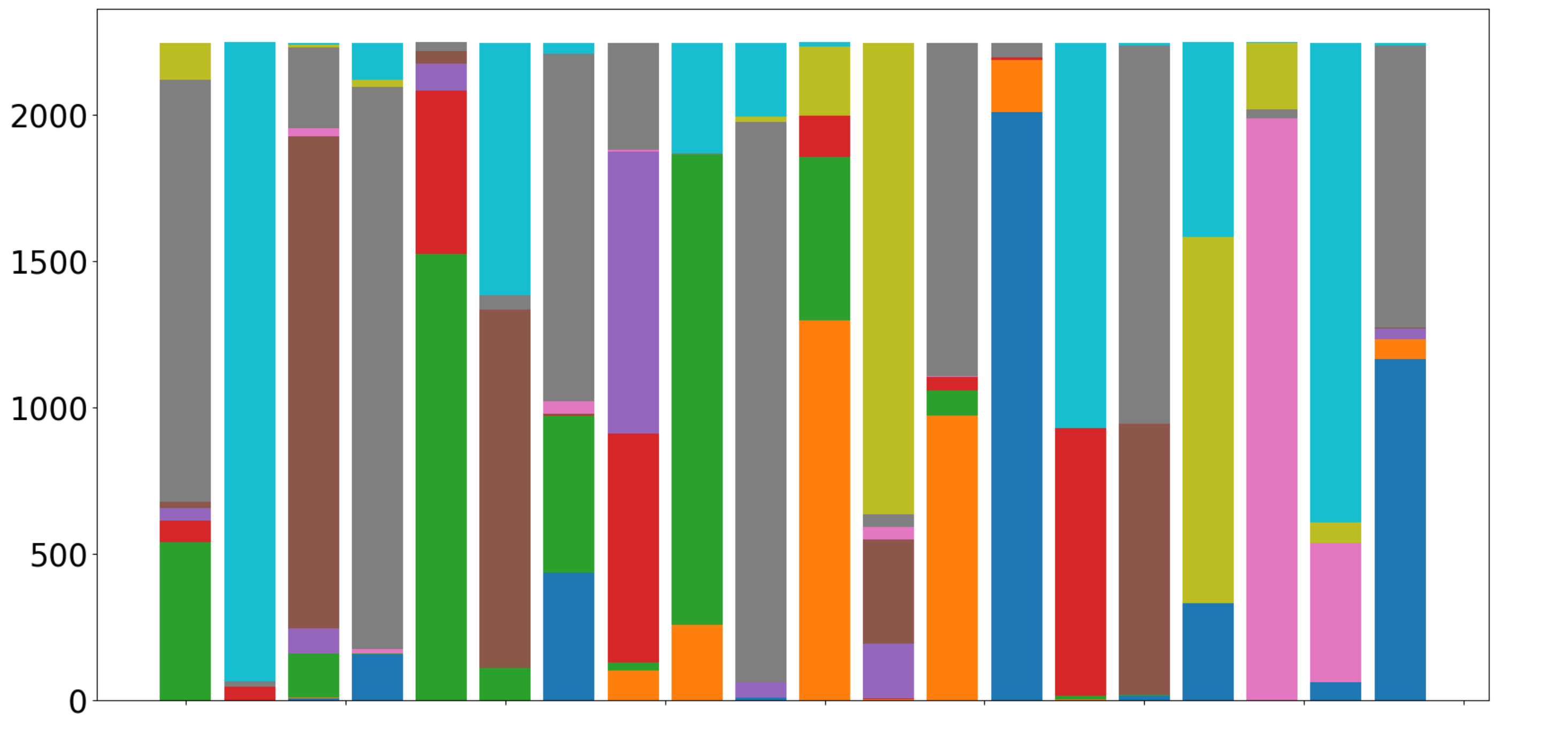}
         \caption{$\alpha=0.1$}
     \end{subfigure}
     \begin{subfigure}[b]{0.32\textwidth}
         \centering
         \includegraphics[width=\textwidth]{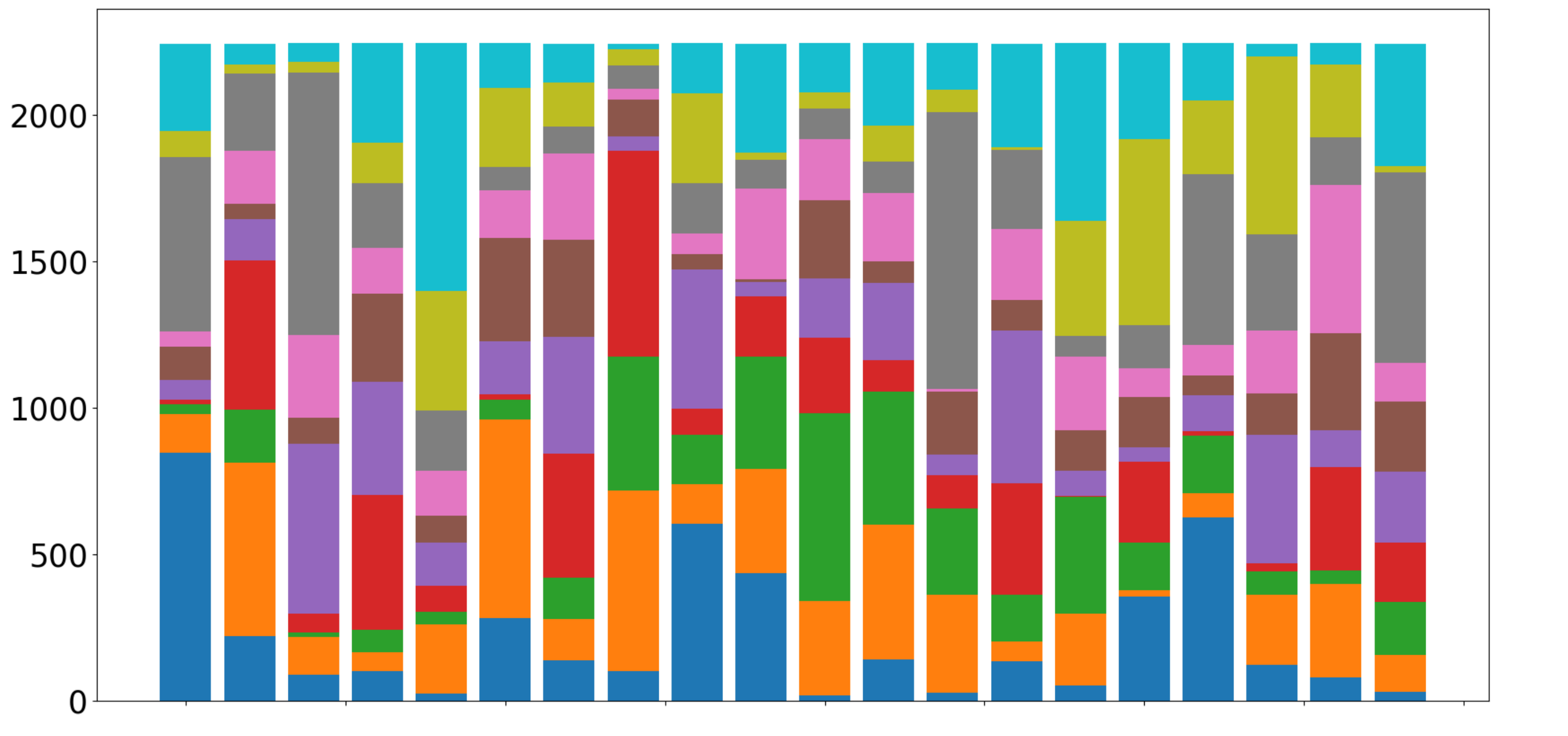}
         \caption{$\alpha=1.0$}
     \end{subfigure}
     \begin{subfigure}[b]{0.32\textwidth}
         \centering
         \includegraphics[width=\textwidth]{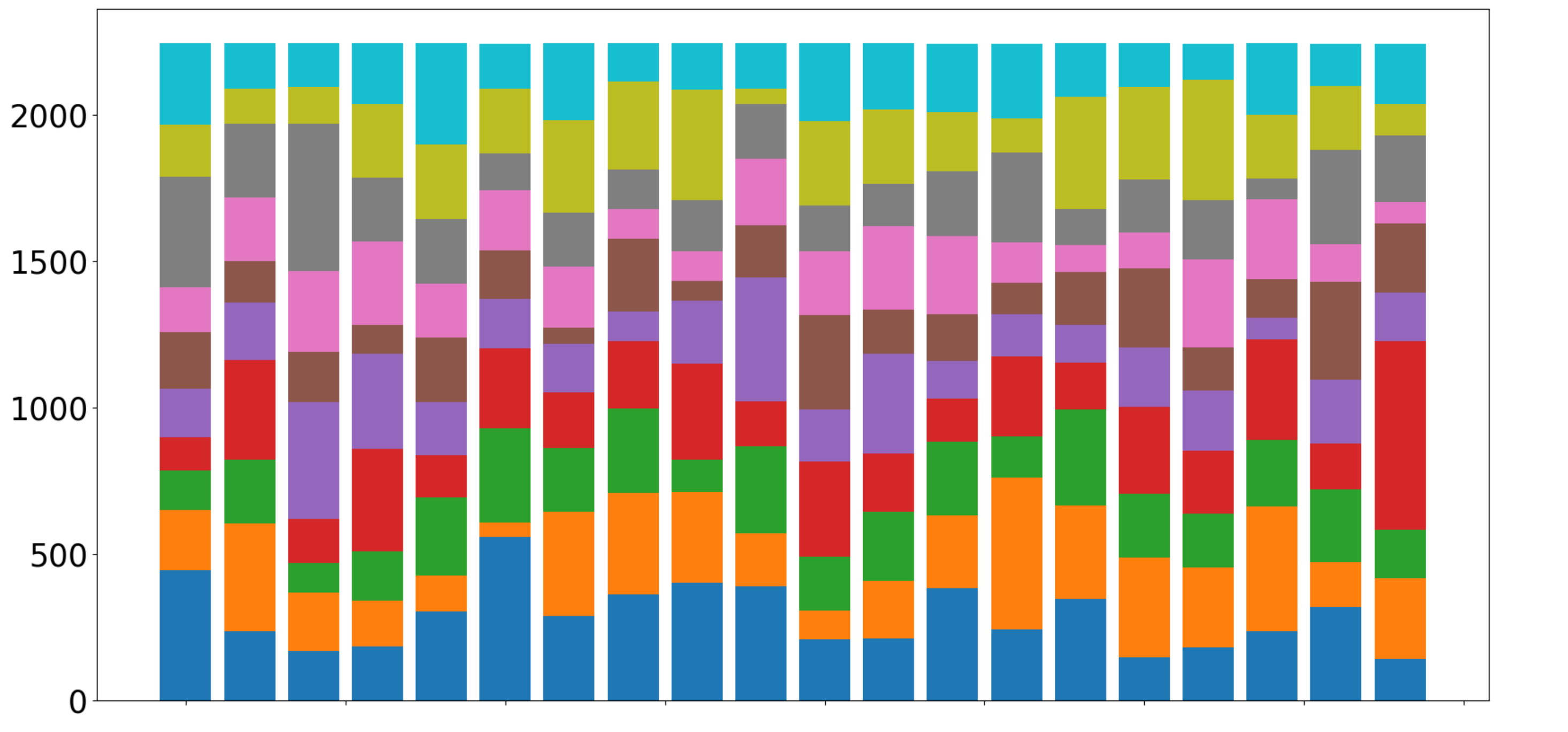}
         \caption{$\alpha=5.0$}
     \end{subfigure}
        \caption{\textbf{Datasets across 20 clients according to IIDness\,($\alpha)$ and heterogeneity\,(top: client, bottom: class)}. The x-axis denotes each client and the y-axis denotes the number of data examples.}
        \label{fig:data_dist}
\end{figure}

{\bf Experimental Setting}. We compared FedCA with three algorithms for CIFAR-10 task\,\citep{krizhevsky2009learning} by following the same state-of-the-art configuration and parameter values as suggested by\,\citet{simonyan2014very}. We employed a 11 layer VGG\,\citep{vgg}, SGD with momentum, weight decay, and standard data augmentation. To simulate a wide range of non-IIDness, we designed representative heterogeneity settings based on widely used techniques\,\citep{yurochkin2019bayesian,hsu2019measuring} as follows\,(\autoref{fig:data_dist}):
\begin{itemize}
\item
    \textbf{Client Heterogeneity}\,\citep{yurochkin2019bayesian}: A dataset is partitioned by following $\mathbf{p}_c \sim Dir_K(\alpha \cdot \vec{1}\,)$ that involves allocating $p_{k,c}$ proportion of data examples for class $c$ to client $k$. 
\item
    \textbf{Class Heterogeneity}\,\citep{hsu2019measuring}: Training examples on every client are drawn independently with class labels following a categorical distribution over N classes. Each instance is drawn with $\mathbf{q} \sim Dir(\alpha \cdot \vec{1}\,)$ from a Dirichlet distribution, where $\alpha>0$ is the concentration parameter controlling IIDness among clients.
\end{itemize}

\vspace{-5pt}
\begin{table}[t]
\caption{\textbf{Top-1 accuracy of FedPdp and FedCA according to class heterogeneity, score function, and $SamplingRatio$}.}
\centering
\vspace{+5pt}
\footnotesize\addtolength{\tabcolsep}{-0pt}
\begin{tabular}{@{}cccccc@{}} 
\toprule
\multirow{2}{*}{Heterogeneity} & \multirow{2}{*}{Algorithm} & \multirow{2}{*}{Score function} & \multicolumn{3}{c}{$SamplingRatio(=|S^t|/|S|)$} \\ \cmidrule(l){4-6} 
 & & & 0.2 & 0.4 & 0.6 \\ \cmidrule(r){1-1}\cmidrule(r){2-2}\cmidrule(r){3-3}\cmidrule(r){4-6}
\multirow{3}{*}{Non-IID$_{(\alpha=0.1)}$} & FedPdp & - & 54.12 $_{\pm3.1}$ & 57.75 $_{\pm3.44}$ & 59.20 $_{\pm2.77}$ \\ \cmidrule(r){2-2}\cmidrule(r){3-3}\cmidrule(r){4-6}
& \multirow{2}{*}{FedCA} & Dirac delta & 52.51 $_{\pm2.04}$ & 61.46 $_{\pm4.78}$ & \textbf{71.11} $_{\pm\textbf{0.11}}$ \\ \cmidrule(r){3-3} \cmidrule(r){4-6} & & Classification loss & \textbf{55.48} $_{\pm\textbf{2.00}}$ & \textbf{65.28} $_{\pm\textbf{1.71}}$ & 69.72 $_{\pm1.22}$ \\
\cmidrule(r){1-1}\cmidrule(r){2-2}\cmidrule(r){3-3}\cmidrule(r){4-6}
\multirow{3}{*}{IID $_{(\alpha=5.0)}$} & FedPdp & - & 82.26 $_{\pm0.43}$ & 82.62 $_{\pm\textbf{0.13}}$ & \textbf{83.08} $_{\pm0.05}$ \\ \cmidrule(r){2-2}\cmidrule(r){3-3}\cmidrule(r){4-6}
& \multirow{2}{*}{FedCA} & Dirac delta & 82.26 $_{\pm0.34}$ & 82.6 $_{\pm0.21}$ & 83.03 $_{\pm\textbf{0.04}}$ \\ \cmidrule(r){3-3} \cmidrule(r){4-6} & & Classification loss & \textbf{82.43} $_{\pm\textbf{0.13}}$ & \textbf{82.71} $_{\pm0.25}$ & 82.91 $_{\pm0.34}$ \\ \bottomrule
\end{tabular}
\label{tab:my-table}
\end{table}

{\bf Results}. Above all, to observe the effects of biased model averaging, we compared FedCA\,(FedPdp + CA) with FedPdp. \autoref{tab:my-table} shows that FedCA consistently outperformed FedPdp, but the variances were comparable. Thus, we can infer that the generalization error mostly comes from the biased model averaging of FedPdp and FedCA helps resolve the challenge.

\begin{figure}[t!]
     \centering
      \begin{subfigure}[b]{0.70\textwidth}
         \centering
         \includegraphics[width=\textwidth]{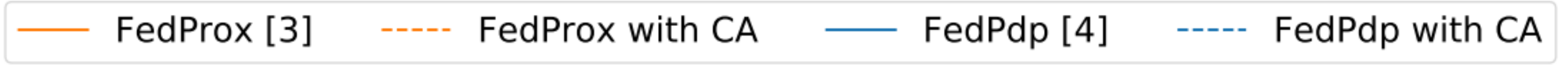}
     \end{subfigure}
     \centering
     \begin{subfigure}[b]{0.44\textwidth}
         \centering
         \includegraphics[width=\textwidth]{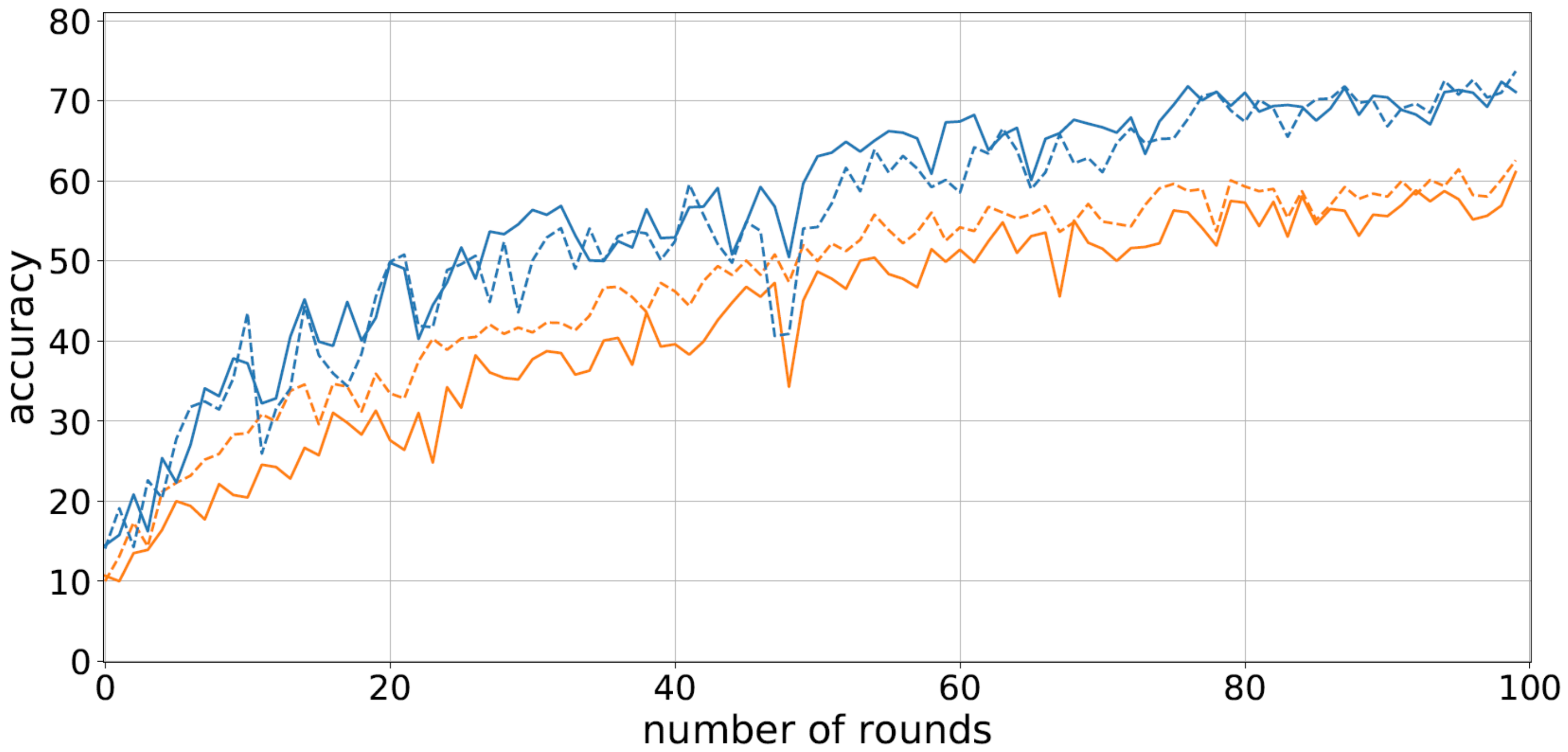}
         \caption{Client Heterogeneity, Dirac delta function}
     \end{subfigure}
     \begin{subfigure}[b]{0.44\textwidth}
         \centering
         \includegraphics[width=\textwidth]{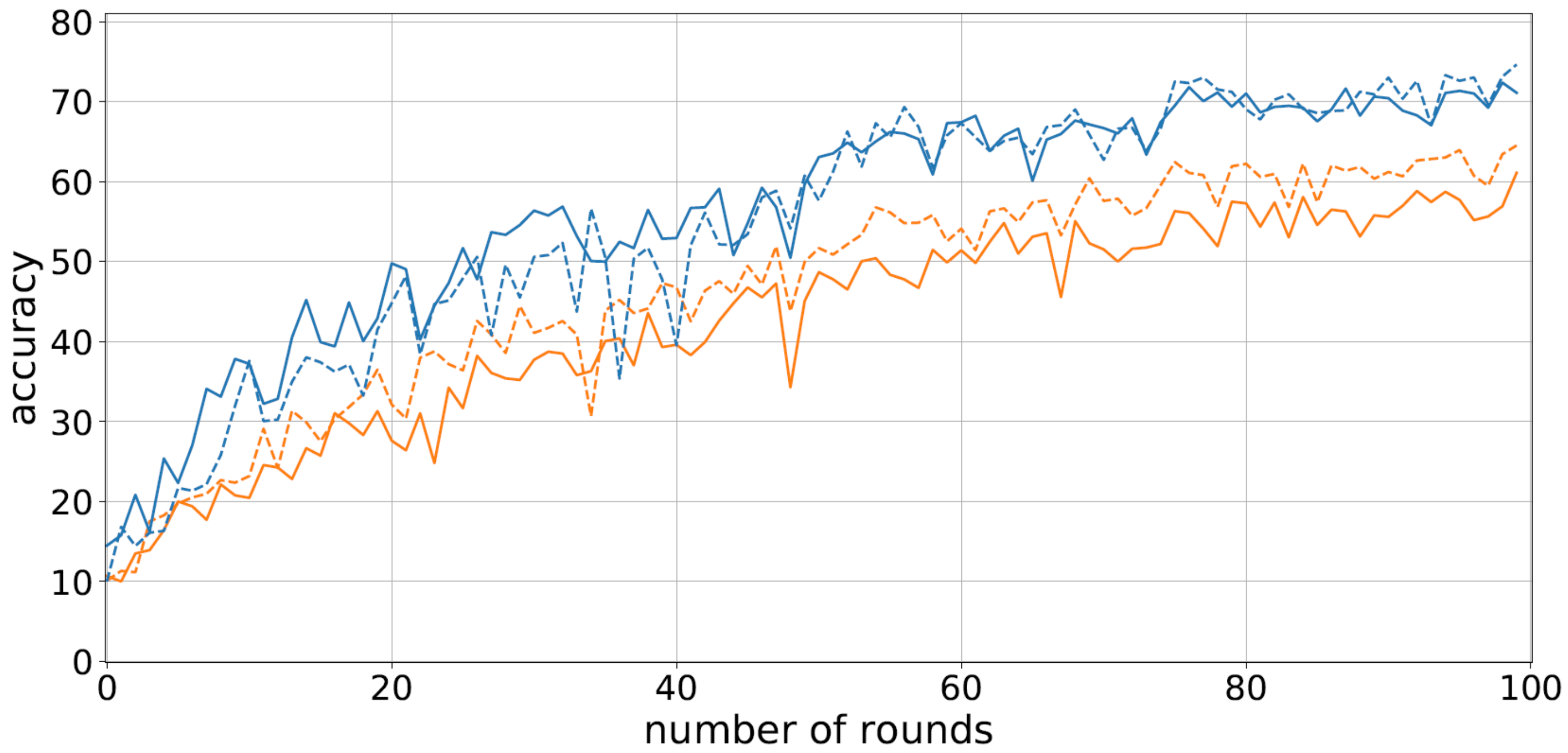}
         \caption{Client Heterogeneity, Classification loss}
     \end{subfigure}
     \centering
     \begin{subfigure}[b]{0.44\textwidth}
         \centering
         \includegraphics[width=\textwidth]{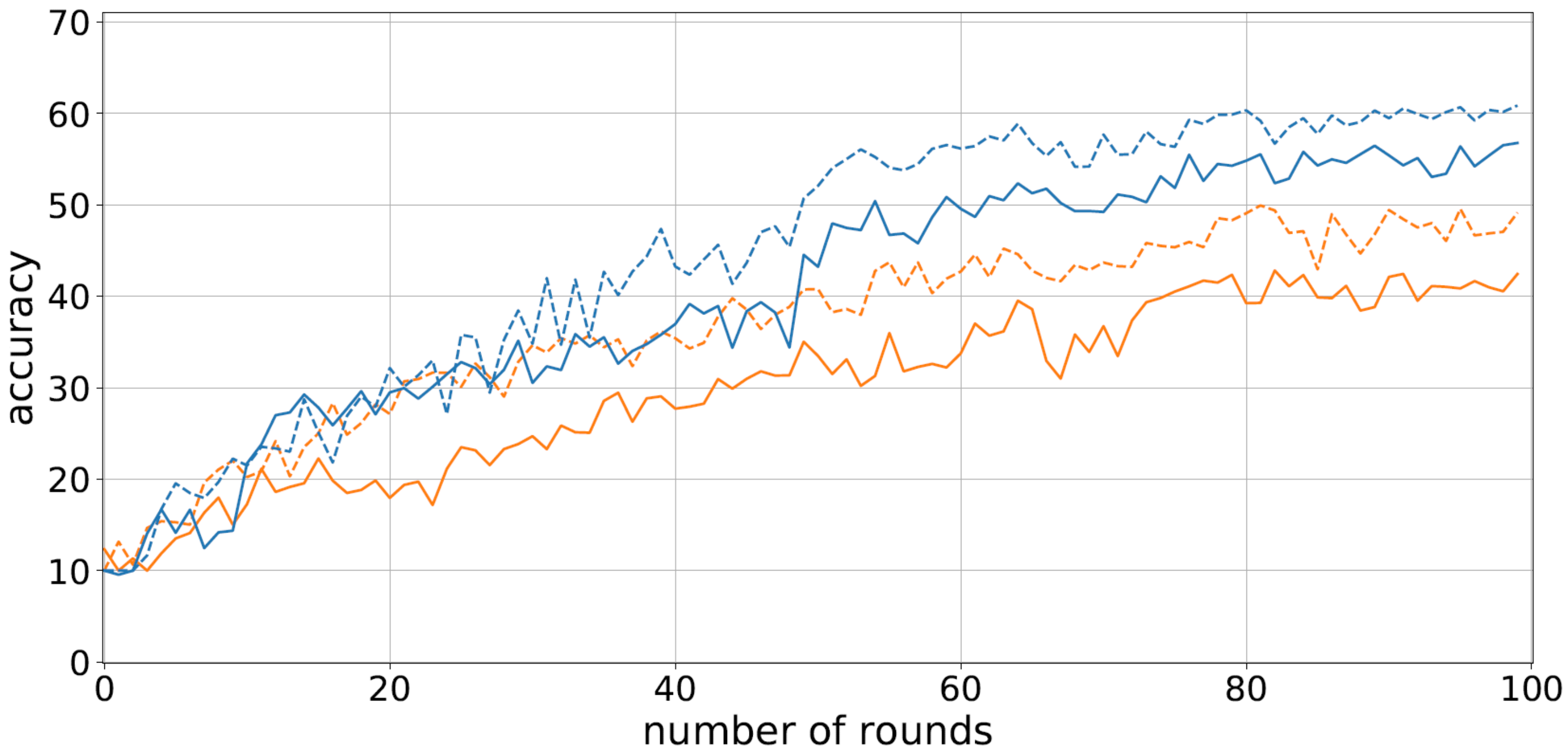}
         \caption{Class Heterogeneity, Dirac delta function}
     \end{subfigure}
     \begin{subfigure}[b]{0.44\textwidth}
         \centering
         \includegraphics[width=\textwidth]{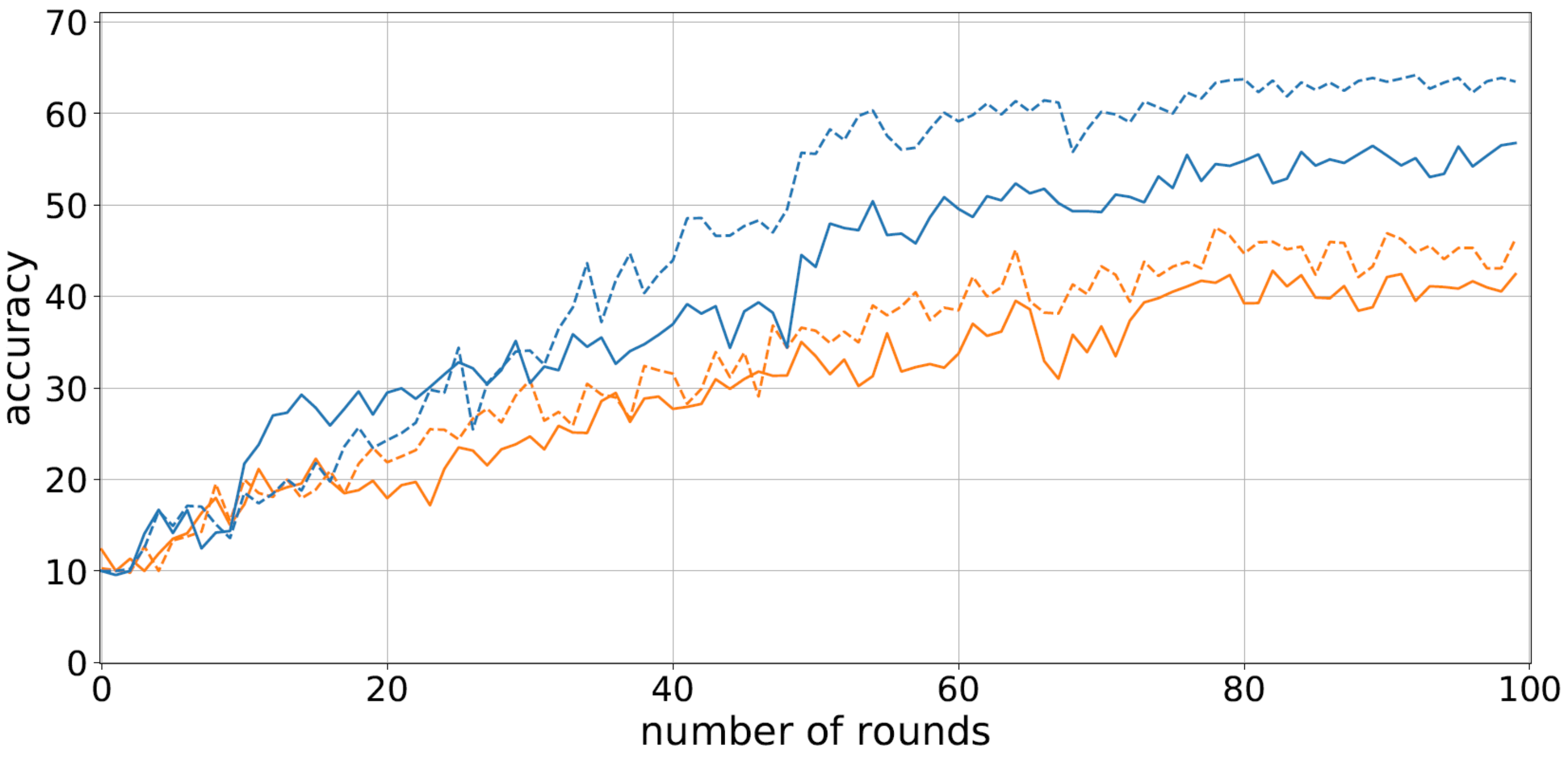}
         \caption{Class Heterogeneity, Classification loss}
     \end{subfigure}

    \caption{\textbf{Effects of Combinatorial Averaging\,(CA)} according to \textbf{heterogeneity}\,(vertical) and \textbf{score function}\,(horizontal). All models are trained with $E=5$ and $SamplingRatio=0.4$ on the non-IID\,($\alpha=0.1$) dataset.}
    \label{fig:CA}
\end{figure}

Furthermore, we compared FedCA with the state-of-the-art algorithms according to different environments such as heterogeneity, score function, and $SamplingRatio$. First, as shown in \autoref{tab:my-table}, FedCA consistently outperformed FedPdp in both IID and non-IID settings. Especially, \autoref{fig:CA} shows that, in both client\,(top) and class\,(bottom) heterogeneity settings, FedProx and FedPdp with CA outperform those without CA in the overall training process. Next, \autoref{fig:CA} also shows that the case of classification loss score\,(right) commonly exhibited less generalization error in various settings than that of Dirac delta score\,(left). Lastly, in the non-IID case of \autoref{tab:my-table}, FedCA facilitated generalization across all sampling ratios and achieved higher accuracy with higher sampling ratio. In particular, when the sampling ratio is $0.6$, FedCA with Dirac delta reached a higher test accuracy by up to $\mathbf{+16.75\%}$. Detailed values in \autoref{fig:CA} is described in \autoref{de-ex-re}.



\section{Multi-Armed Bandit (MAB) based Client Sampling}
\label{sec:MAB}
In this section, we present our novel algorithm, which is coined as \textbf{\textit{FedCM}}(\underline{Fed}erated learning with \emph{\underline{C}ombinatorial averaging} and \emph{\underline{M}AB based client sampling}). Here, for the experiment, we used the same recipe mentioned in Section \ref{sec:CA}.

{\bf Proposed Algorithm: FedCM}. To resolve the challenge of lack of prior knowledge in client sampling, we introduce a MAB based client sampling scheme to reflect prior knowledge that models previous client sampling behavior. Unlike the conventional schemes, as compared in \autoref{tab:algorithm}, MAB based client sampling can incorporate prior knowledge by prioritizing the clients that were subsampled in the last iteration.
By integrating the combinatorial averaging of FedCA with MAB based client sampling, we could simultaneously resolve both aforementioned challenges, and we call this integrated extension FedCM that extends $SampleClients$ and $PriorKnowledge$ of Algorithm\,\ref{al:Framework} as follows. In Algorithm\,\ref{al:Framework}, the function $SampleClients$ takes the total client set $S$, $SamplingRatio$, and $PriorKnowledge$ as input, and returns the sampled client set $S^t$ as output. Based on what $PriorKnowledge$ is provided as well as how $SampleClients$ handles the $PriorKnowledge$, FedCM is derived into two heuristic algorithms with regard to the framework of representative MAB algorithms such as UCB\,(Upper Confidence Bound)\,\citep{combucb} and TS\,(Thompson sampling)\,\citep{combts}: \emph{FedCM-UCB} and \emph{FedCM-TS}.

\begin{algorithm}[H]
\label{al:ucb}
\SetAlgoLined
\SetKwInOut{Input}{\textsc{Input}}
\SetKwInOut{Output}{\textsc{Output}}
\caption{FedCM-UCB\,(Upper Confidence Bound)}
\label{al:FedCM-UCB}
\Input{$S, SamplingRatio, \mathcal{P} = (\mathcal{P}_1, \dots ,\mathcal{P}_n)$ where $\mathcal{P}_n = (\hat{\mu}_n, t, a_n), \mathcal{F}_{t-1}$}
\Output{$S^t$}
\For{each client $k \in S^{t-1}$}{
    Update $(\hat{\mu}_k, a_k) \leftarrow ((a_k \hat{\mu}_k + r_{k}) \,/ \,(a_k + 1), \, a_{k} + 1$) where $r_{k} = \mathbf{1}_{[{k} \in S^{t-1}_{opt}]}$ \\
    Set $\bar{\mu}_k \leftarrow \hat{\mu}_k + \sqrt{\frac{3 \ln{t}}{2 a_t}}$
}
$S^t \leftarrow \{ n | \, \forall n \in S, \, |\{ m | \, \forall m \in S, \, \bar{\mu}_m \geq \bar{\mu}_n\}|/|S| \leq SamplingRatio \}$ \\
\end{algorithm}

First, Algorithm\,\ref{al:FedCM-UCB} shows how FedCM extends $SampleClients$ to UCB. $PriorKnowledge$ of FedCM-UCB involves prior knowledge $\mathcal{P}$ and $\sigma$-field $\mathcal{F}_{t-1}$ generated by the previous observations $S^1, S^1_{opt}, \ldots, S^{t-1}, S^{t-1}_{opt}$. FedCM-UCB iteratively updates $\hat{\mu}_k$ of each client with reward $r_{k} = \mathbf{1}_{[{k} \in S^{t-1}_{opt}]}$\,(Lines 1--3) and samples clients based on $\bar{\mu}_k$\,(Line 4). 

\begin{algorithm}[H]
\label{al:ts}
\SetAlgoLined
\SetKwInOut{Input}{\textsc{Input}}
\SetKwInOut{Output}{\textsc{Output}}
\caption{FedCM-TS\,(Thompson Sampling)}
\label{al:FedCM-TS}
\Input{$S, SamplingRatio, \mathcal{P} = (\mathcal{P}_1, \dots ,\mathcal{P}_n)$ where $\mathcal{P}_n \sim Beta(\alpha_n, \beta_n), \mathcal{F}_{t-1}$}
\Output{$S^t$}
\For{each client $k \in S^{t-1}$}{
    Update $(\alpha_{k}, \beta_{k}) \leftarrow (\alpha_{k} + r_{k}, \beta_{k} + 1 - r_{k}$) where $r_{k} = \mathbf{1}_{[{k} \in S^{t-1}_{opt}]}$ \\
    Draw a sample $\hat{\theta}_k$ according to $\mathcal{P}_k$ 
}
$S^t \leftarrow \{ n | \, \forall n \in S, \, |\{ m | \, \forall m \in S, \, \hat{\theta}_m \geq \hat{\theta}_n\}/|S| \leq SamplingRatio \}$ \\
\end{algorithm}

Next, Algorithm\,\ref{al:FedCM-TS} shows how FedCM extends $SampleClients$ to TS, which is one of the most promising algorithms in bandit problems. $PriorKnowledge$ of FedCM-TS involves a beta distribution and the same $\sigma$-field $\mathcal{F}_{t-1}$ as the one of FedCM-UCB. FedCM-TS iteratively updates $\alpha_{k}, \beta_{k}$ of each client with reward $r_{k} = \mathbf{1}_{[{k} \in S^{t-1}_{opt}]}$\,(Lines 1--3) and samples clients based on $\hat{\theta}_k$\,(Line 4). Detailed settings are further illustrated in \autoref{sec:bandit}

\begin{figure}[t]
     \centering
     \begin{subfigure}[b]{0.90\textwidth}
         \centering
         \includegraphics[width=\textwidth]{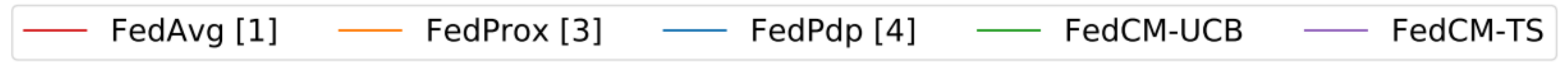}
         \vspace{-10pt}
     \end{subfigure}
     \centering
     \begin{subfigure}[b]{0.475\textwidth}
         \centering
         \includegraphics[width=\textwidth]{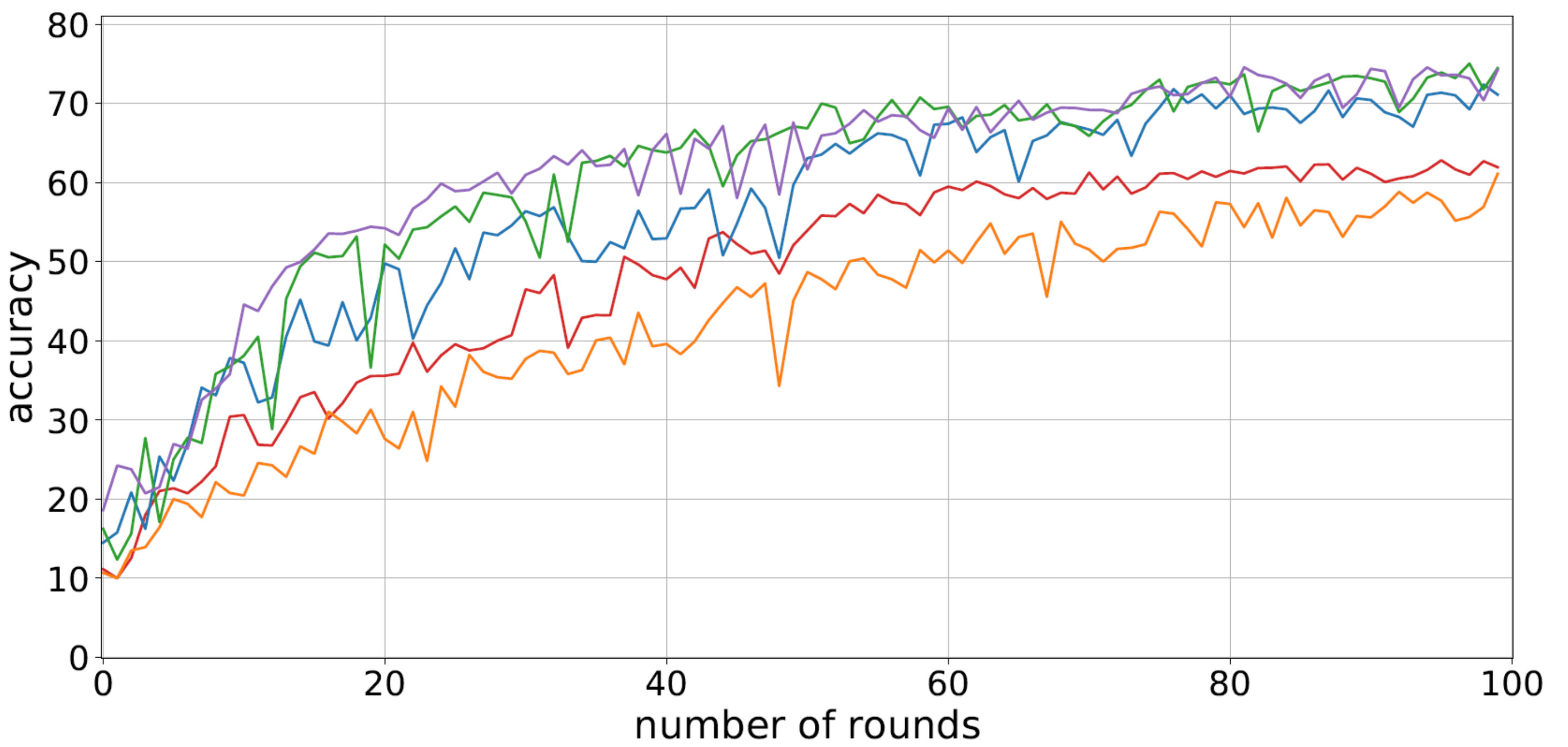}
         \caption{Client Heterogeneity, Dirac delta function}
     \end{subfigure}
     \begin{subfigure}[b]{0.475\textwidth}
         \centering
         \includegraphics[width=\textwidth]{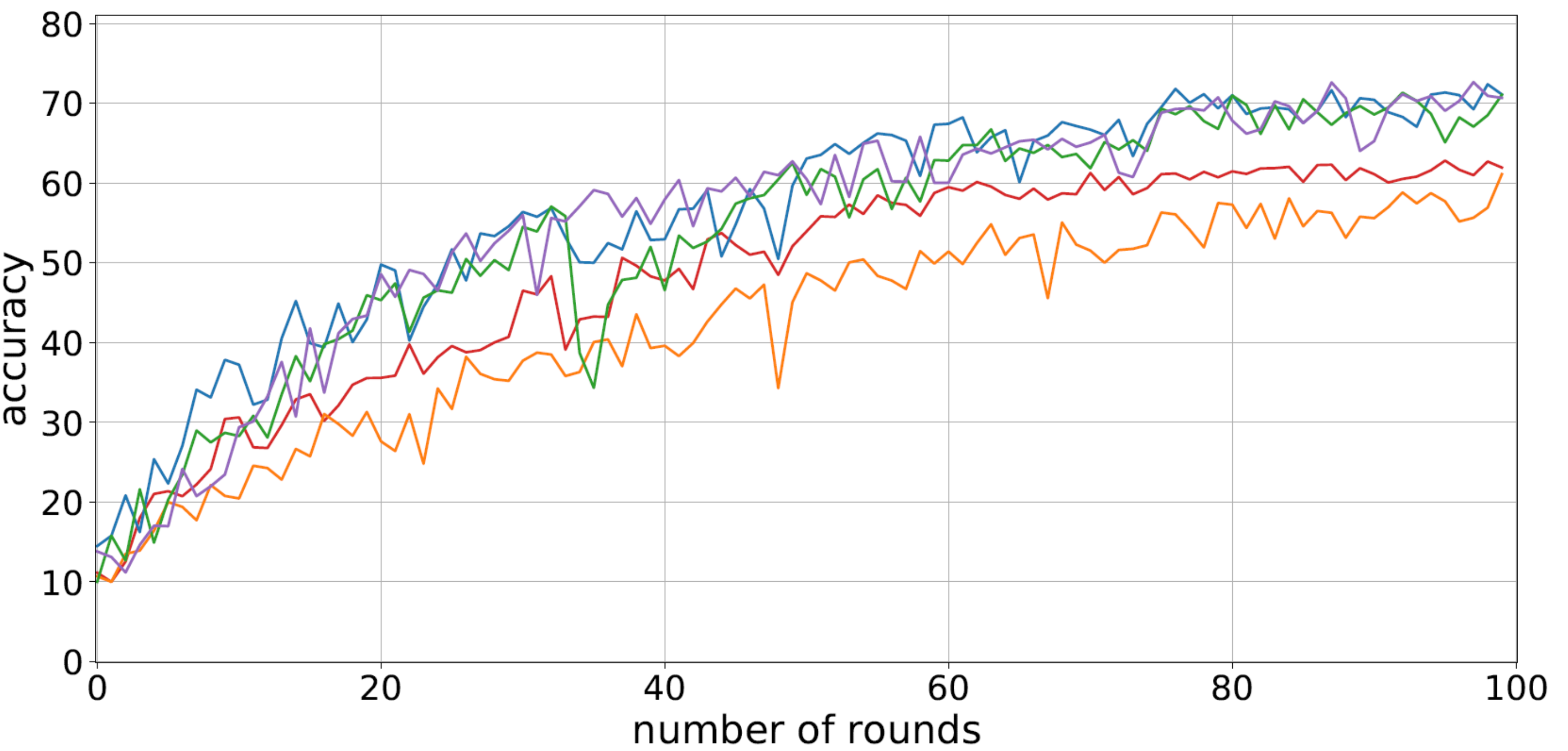}
         \caption{Client Heterogeneity, Classification loss}
     \end{subfigure}
     \centering
     \begin{subfigure}[b]{0.475\textwidth}
         \centering
         \includegraphics[width=\textwidth]{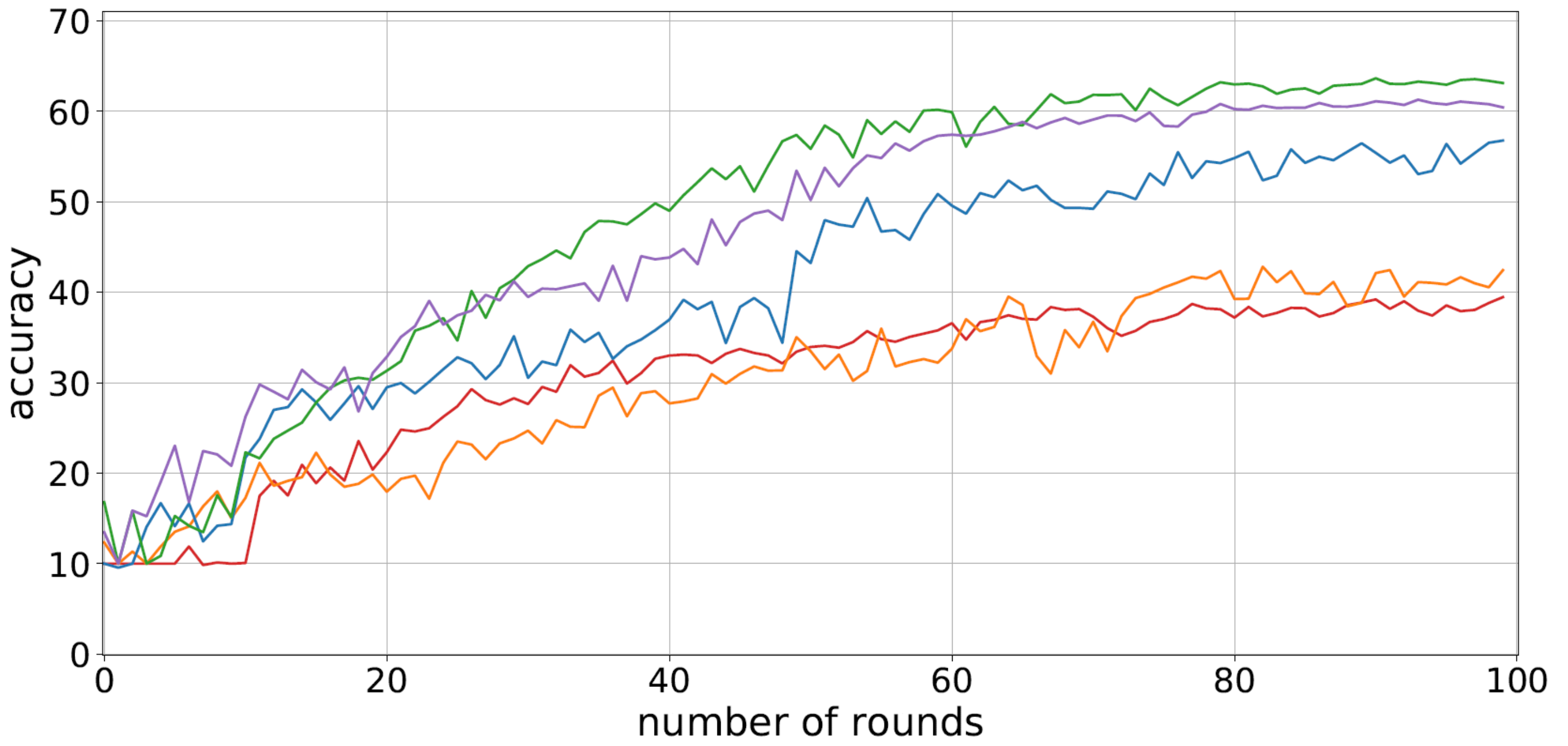}
         \caption{Class Heterogeneity, Dirac delta function}
     \end{subfigure}
     \begin{subfigure}[b]{0.475\textwidth}
         \centering
         \includegraphics[width=\textwidth]{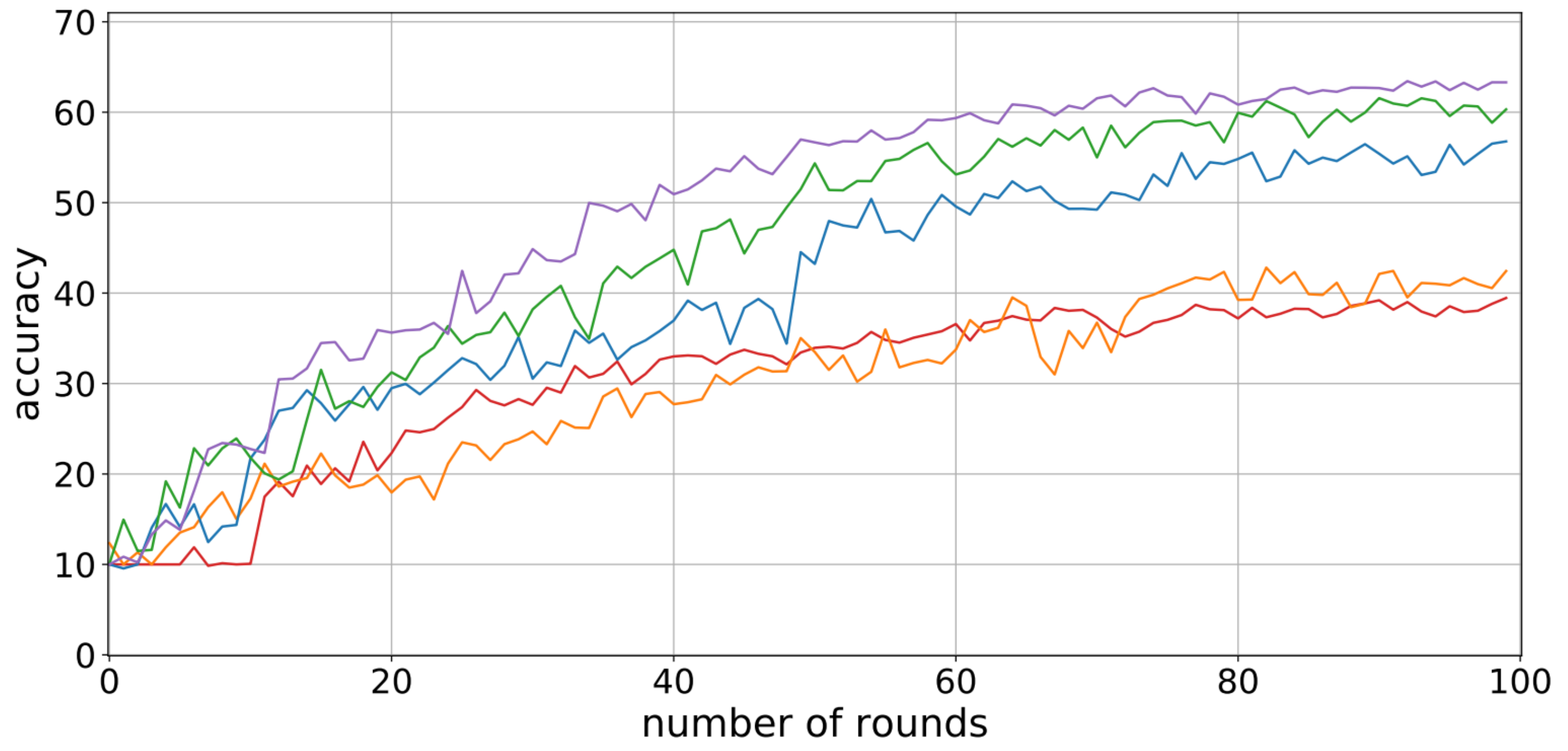}
         \caption{Class Heterogeneity, Classification loss}
     \end{subfigure}
     
    \caption{\textbf{Effects of FedCM} according to \textbf{heterogeneity}\,(vertical) and \textbf{score function}\,(horizontal). All models are trained with $E=5$ and $SamplingRatio=0.4$ on the non-IID\,($\alpha=0.1$) dataset.}
    \label{fig:MAB}
\end{figure}

{\bf Results}. 
Despite the improvement through FedCA, it was observed that the convergence speed becomes significantly slower in non-IID settings than in IID settings\,(see \autoref{fig:conver_spe} in \autoref{de-ex-re}).
We systematically compared FedCM with the state-of-the-art algorithms according to class heterogenity and score function. As described in \autoref{fig:MAB}, interestingly, both FedCM-UCB and Fed-TS outperformed FedAvg and FedProx in all cases and FedPdp in the case of class heterogeneity in terms of both generalization error and convergence speed. These improvements occured throughout the whole training processes, even in the early stage of training. Detailed values of the accuracies and convergence speed are in \autoref{de-ex-re}.

\vspace{-10pt}
\section{Related Work}
\vspace{-5pt}

Recent studies \citep{fedprox, li2019convergence} emphasizes in-depth investigation of client sampling and model averaging. FedCS\,\citep{nishio2019client} aims at maximizing the number of clients while minimizing the overall communication delay for a set of sampled learners by considering a round-trip time constraint. \citet{agnosticfl} optimizes the degree of client participation via a fairness objective function that enables the model to be agnostic to any mixture of client data distribution. In\,\citet{powerofchoice}, an optimal set is subsampled from the sampled clients based on the loss of each client's local data. Their concept is quite similar while it differs from ours in that the criterion for the optimality changes to the training loss from the validation loss and they did not consider any prior knowledge. In addition, it may be truly suboptimal because in the course of sampling, the local clients that penalized the training with adverse effects in the past are not considered at all. On the other side, for the purpose of communication reduction, reinforcement learning\citep{nadiger2019federated,rlclientsampling,zhuo2019federated} and MAB \citep{mabclientsampling,xia2020multi} algorithms are being widely investigated.


\section{Conclusion}
In this paper, we formulated a novel system-level framework of \emph{FL with knowledgeable sampling and filtered averaging} to address the challenge of biased model averaging and lack of prior knowledge in client sampling. To this end, we presented our novel algorithm called \textbf{\textit{FedCM}} that resolves the two challenges by filtering biased models with combinatorial averaging and utilizing prior knowledge with multi-armed bandit based client sampling. Interestingly, combinatorial averaging itself significantly improved the performance of conventional algorithms, and the application of both techniques led to greater synergy. Experimental results show that, compared with the state-of-the-art algorithms, FedCM improved the test accuracy by up to $\mathbf{37.25\%}$ and convergence rate by up to $\mathbf{4.17}$ times.

\bibliography{main.bib}
\bibliographystyle{unsrtnat}

\appendix
\section{Overview of FedCA}
\label{app:fedca}

\begin{figure}[h!]
     \centering
     \includegraphics[width=\textwidth]{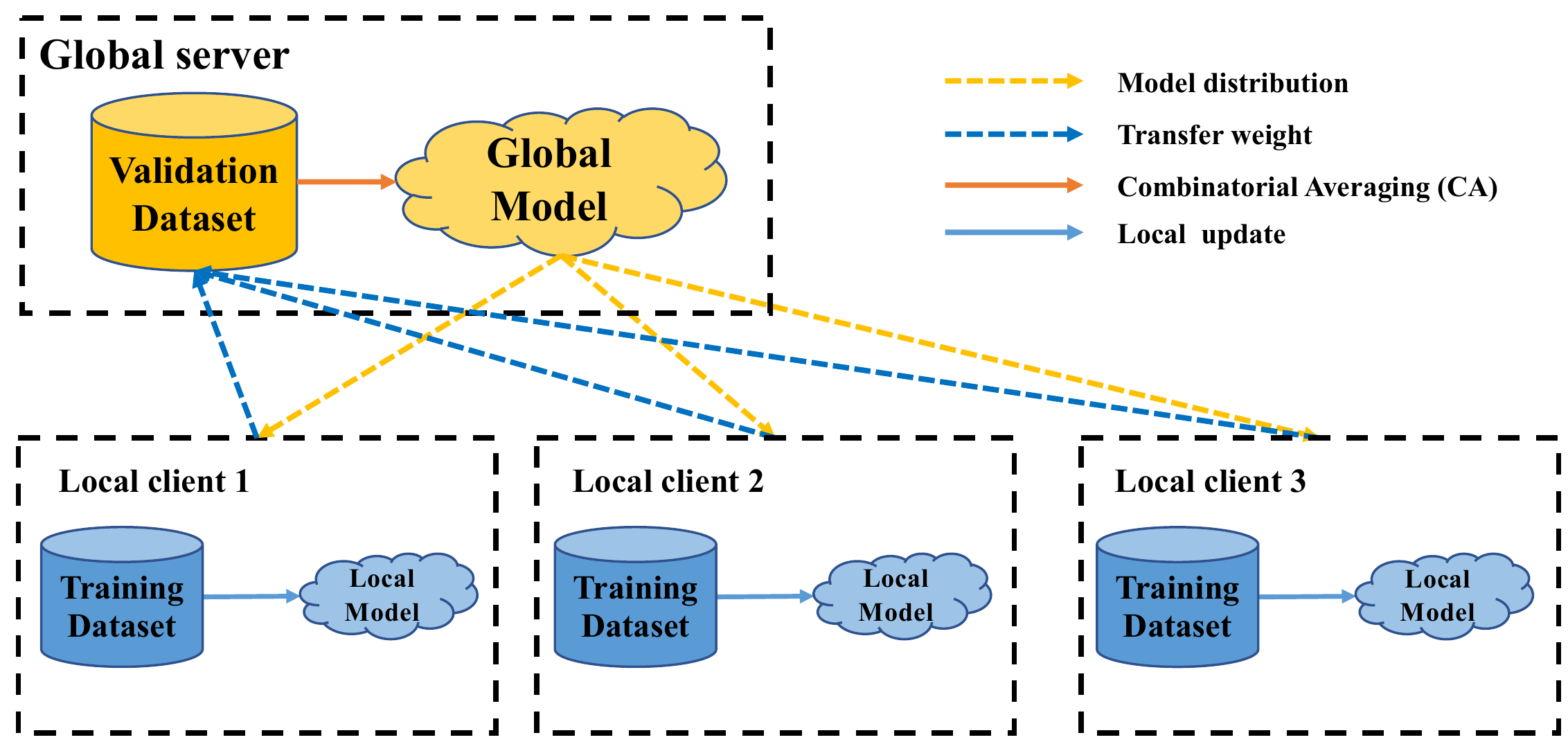}
     \vspace{-10pt}
     \caption{\textbf{An overview of FedCA.} The dashed lines connected between a global server and local clients is related to the communication cost, and the solid line within the server and clients means the computing cost.}
    \label{fig:Main Flow}
\end{figure}

\paragraph{Overview.}
An overview of FedCA in the \autoref{sec:CA} is summaraized in \autoref{fig:Main Flow}. An apparent difference between the existing FL framwork and ours is the existence of the validation dataset in the Global server. Similar to FedPdp\,\cite{li2019convergence}, after partial device participation, each participant transfers locally updated weights to Global Model. By the way, in our framework, global server subsamples the optimal combinations of clients as described in \autoref{eq:CombOpt}. After this combinatorial optimization, the subsampled updates are aggregated and distributed into each corresponding local client.

As above, CA does not have influences on the sampling scheme, so it can have a plug-and-play nature, i.e., CA can be incorporated with uniform sampling \citep{fedavg, fedprox, li2019convergence} or multi-armed bandit based sampling.



\section{Client Variance}
\begin{figure}[h!]
     \centering
     \begin{subfigure}[b]{0.90\textwidth}
         \centering
         \includegraphics[width=\textwidth]{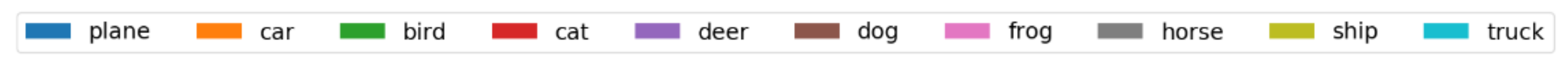}
         \vspace{-10pt}
     \end{subfigure}
     \centering
     \begin{subfigure}[b]{0.24\textwidth}
         \centering
         \includegraphics[width=\textwidth]{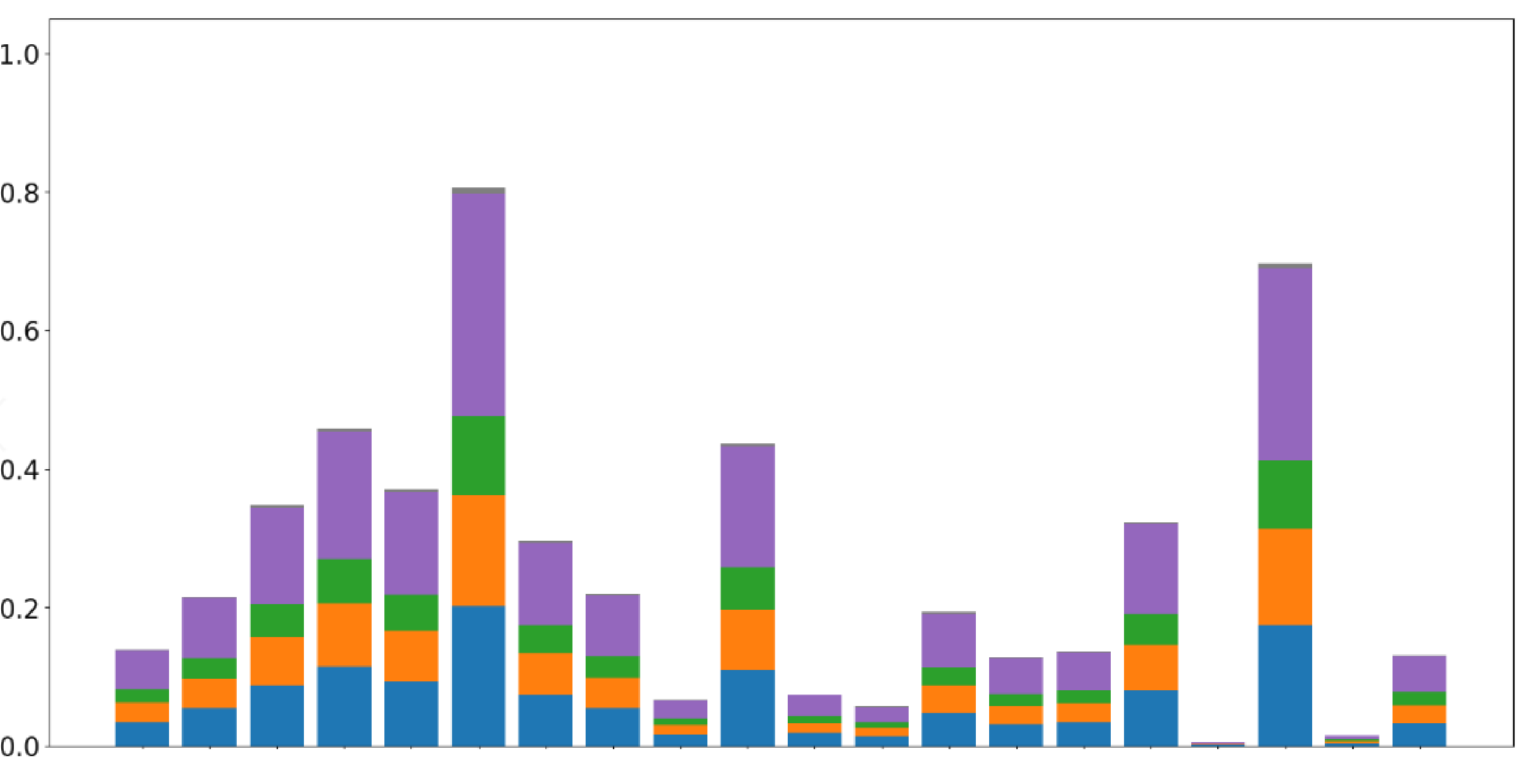}
         \caption{$\alpha=0.1, E=1$}
     \end{subfigure}
     \hfill
     \begin{subfigure}[b]{0.24\textwidth}
         \centering
         \includegraphics[width=\textwidth]{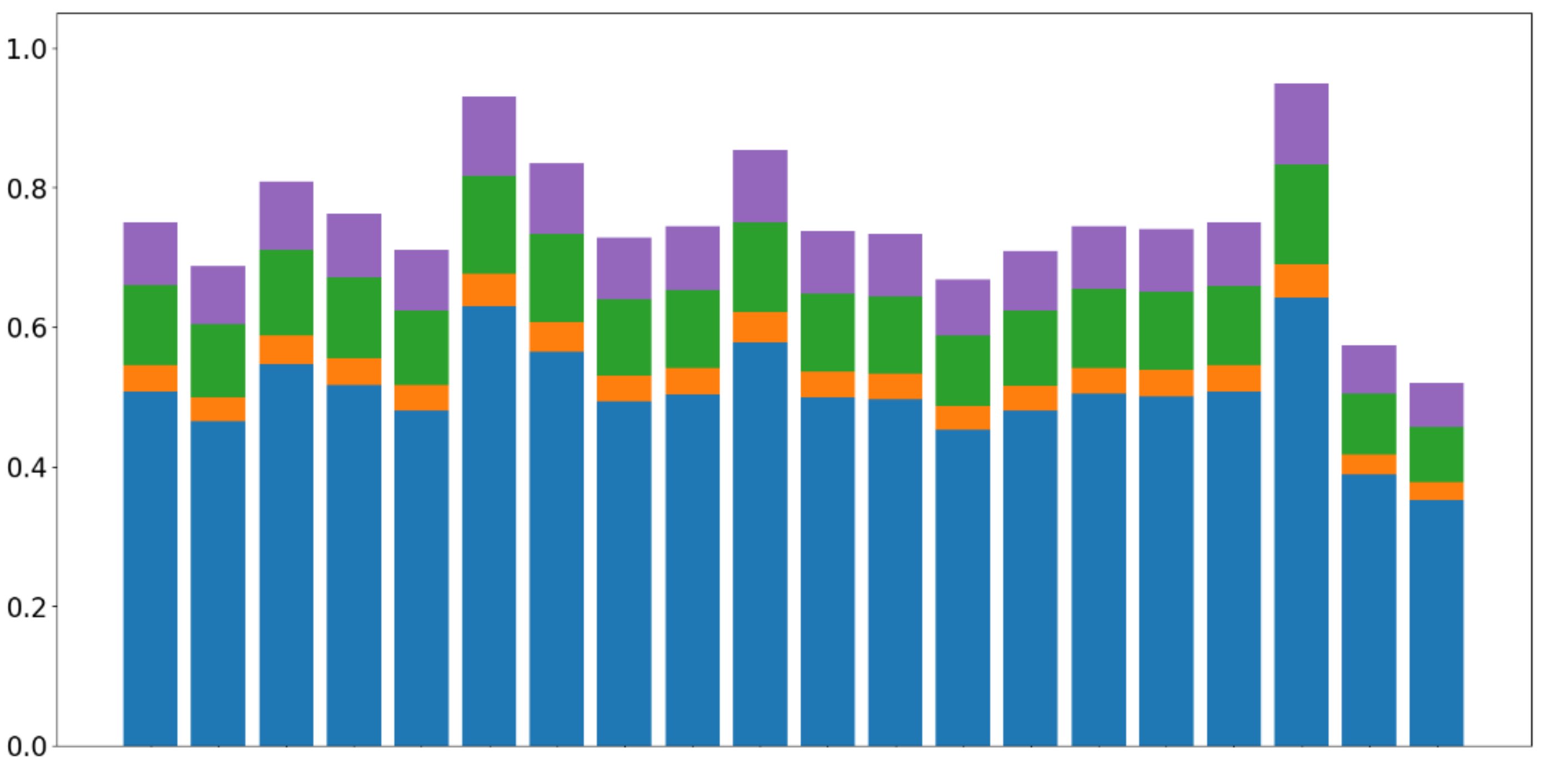}
         \caption{$\alpha=0.1, E=5$}
     \end{subfigure}
     \hfill
     \begin{subfigure}[b]{0.24\textwidth}
         \centering
         \includegraphics[width=\textwidth]{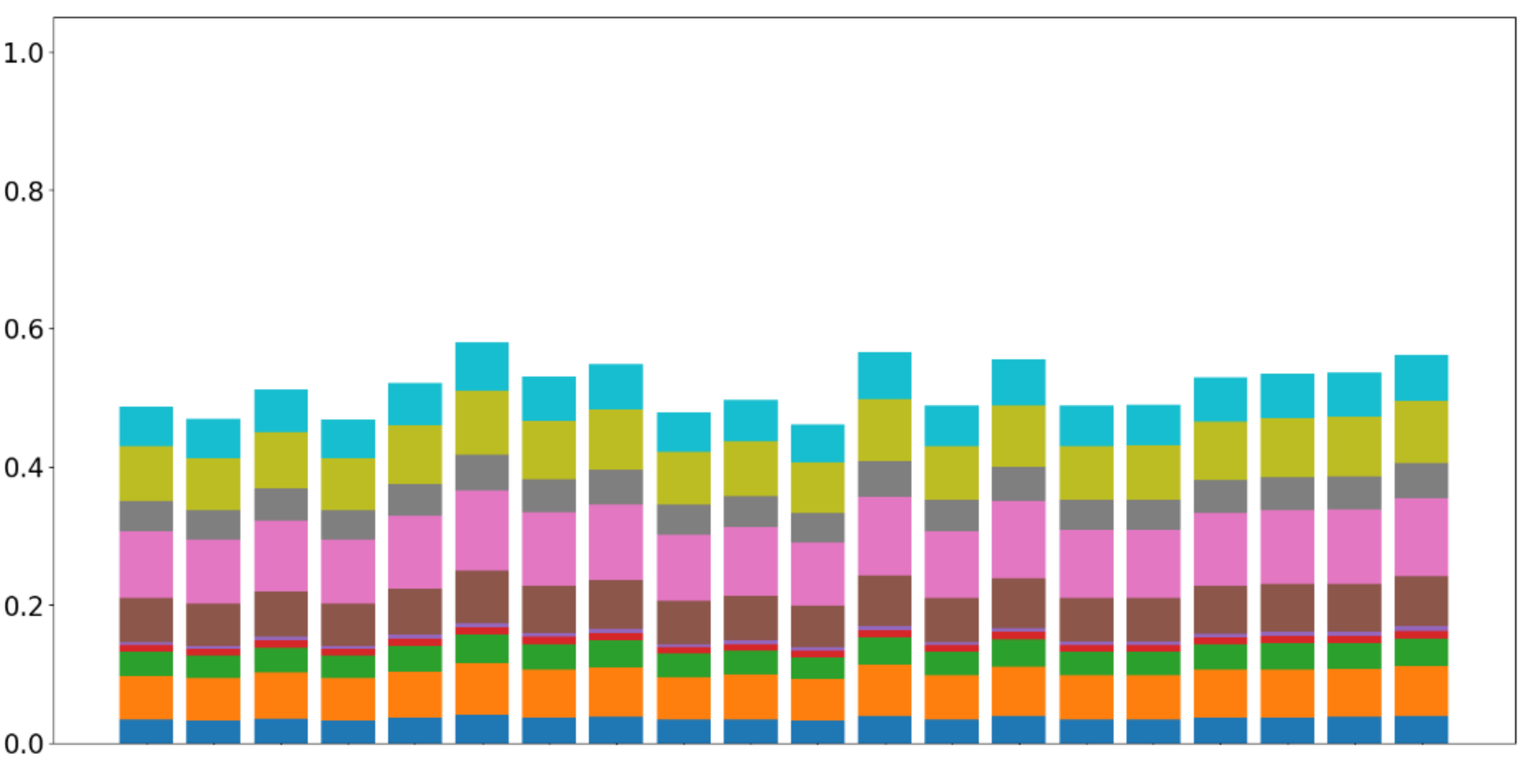}
         \caption{$\alpha=5.0, E=1$}
     \end{subfigure}
     \hfill
     \begin{subfigure}[b]{0.24\textwidth}
         \centering
         \includegraphics[width=\textwidth]{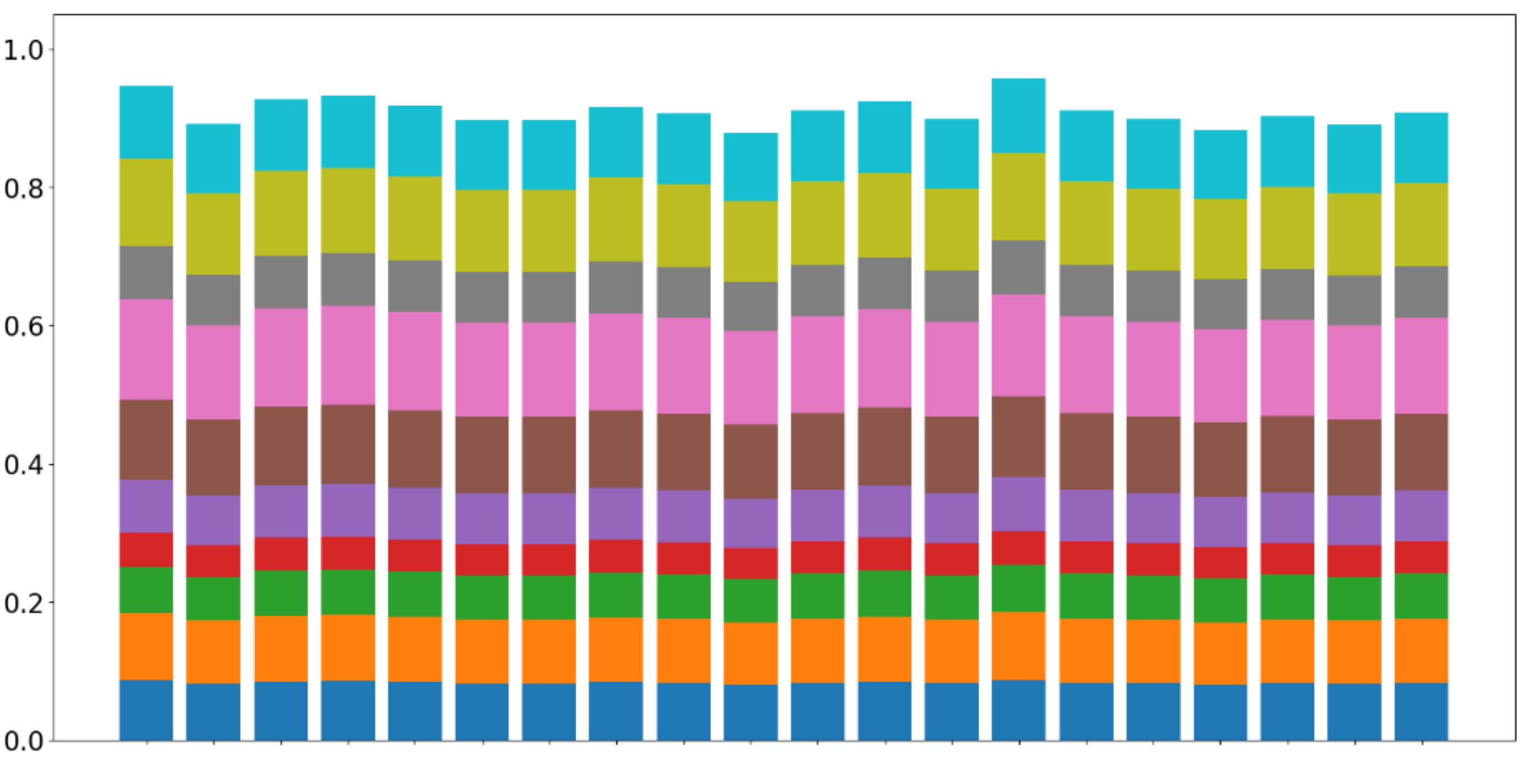}
         \caption{$\alpha=5.0, E=5$}
     \end{subfigure}
     \centering
     \begin{subfigure}[b]{0.24\textwidth}
         \centering
         \includegraphics[width=\textwidth]{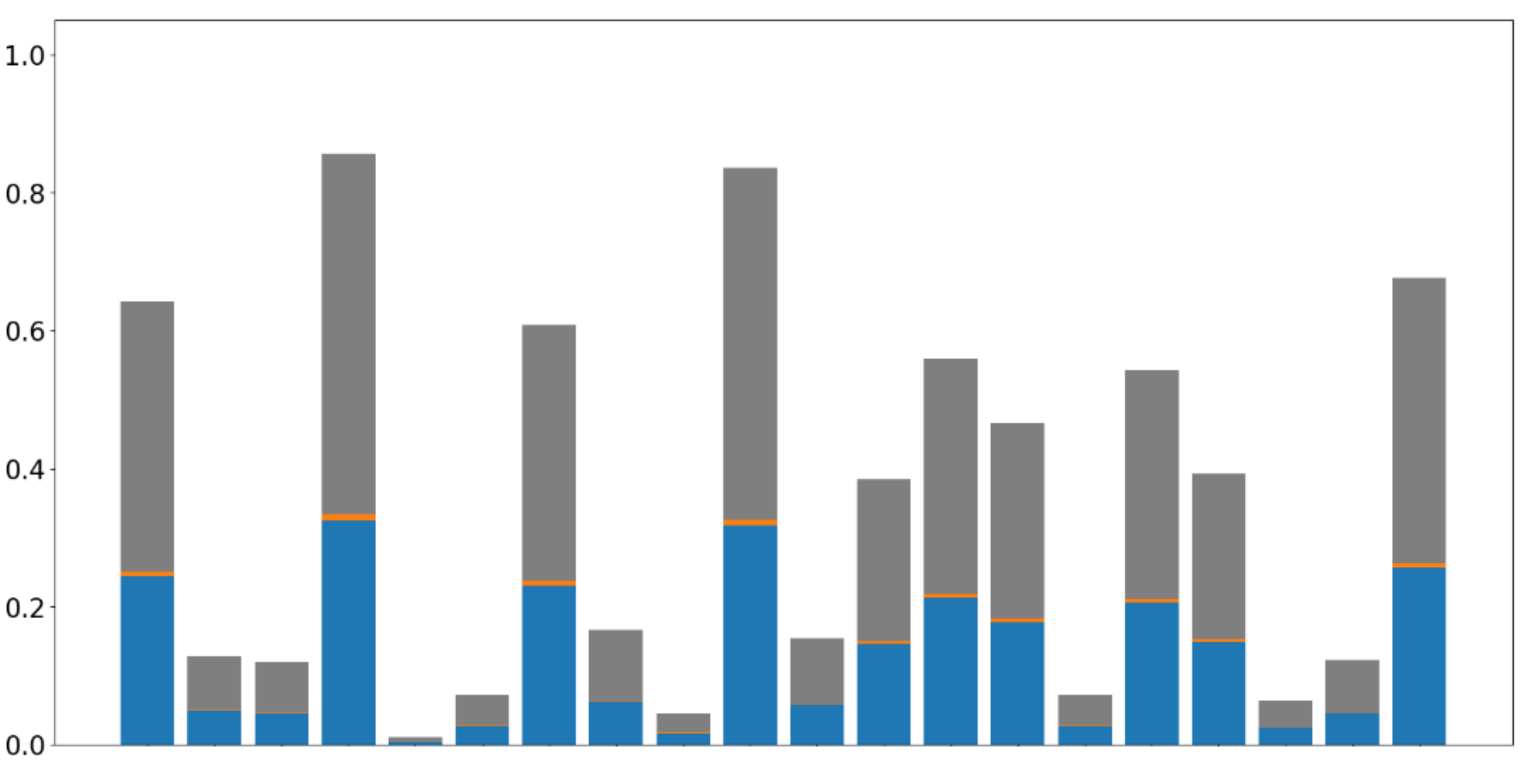}
         \caption{$\alpha=0.1, E=1$}
     \end{subfigure}
     \hfill
     \begin{subfigure}[b]{0.24\textwidth}
         \centering
         \includegraphics[width=\textwidth]{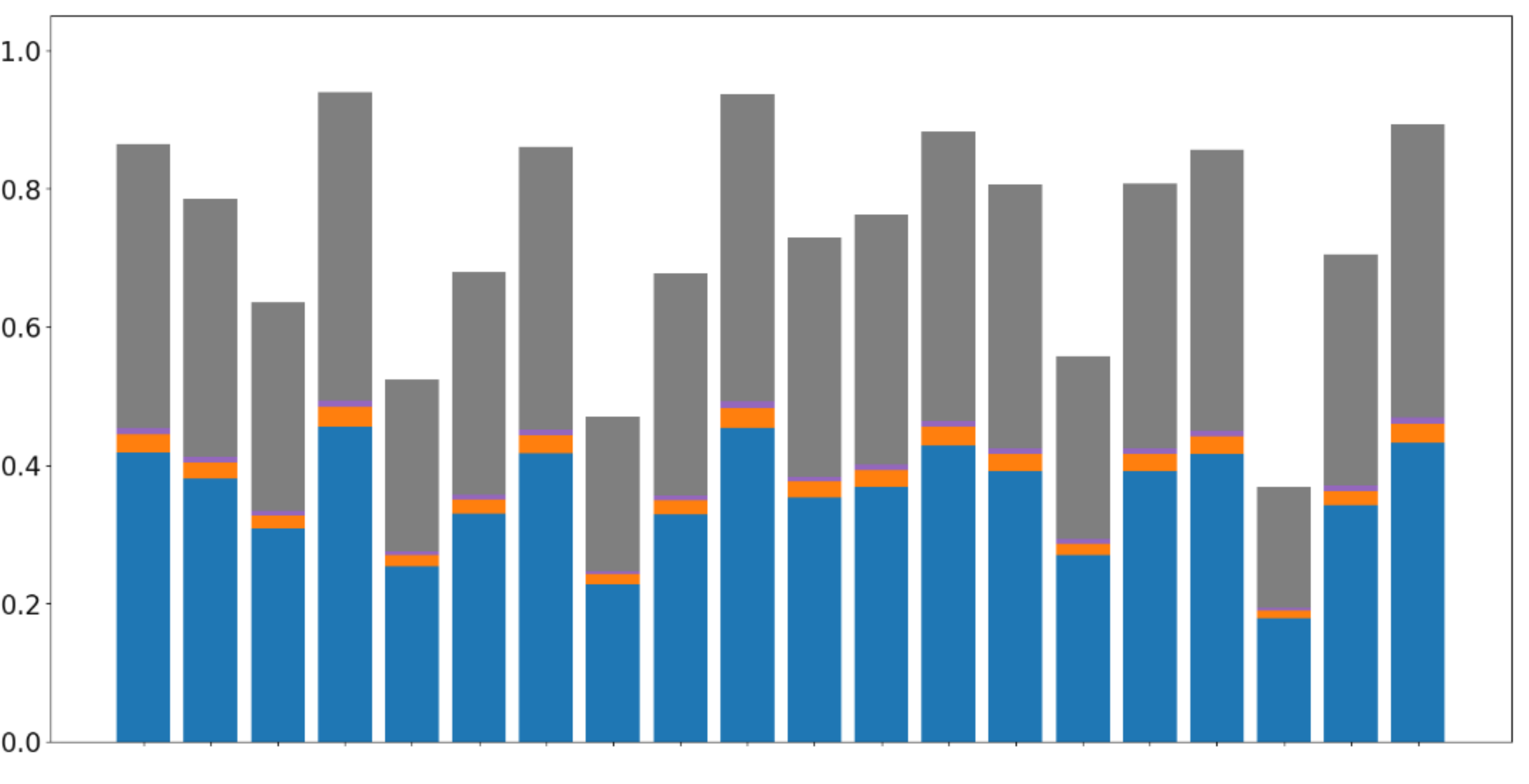}
         \caption{$\alpha=0.1, E=5$}
     \end{subfigure}
     \hfill
     \begin{subfigure}[b]{0.24\textwidth}
         \centering
         \includegraphics[width=\textwidth]{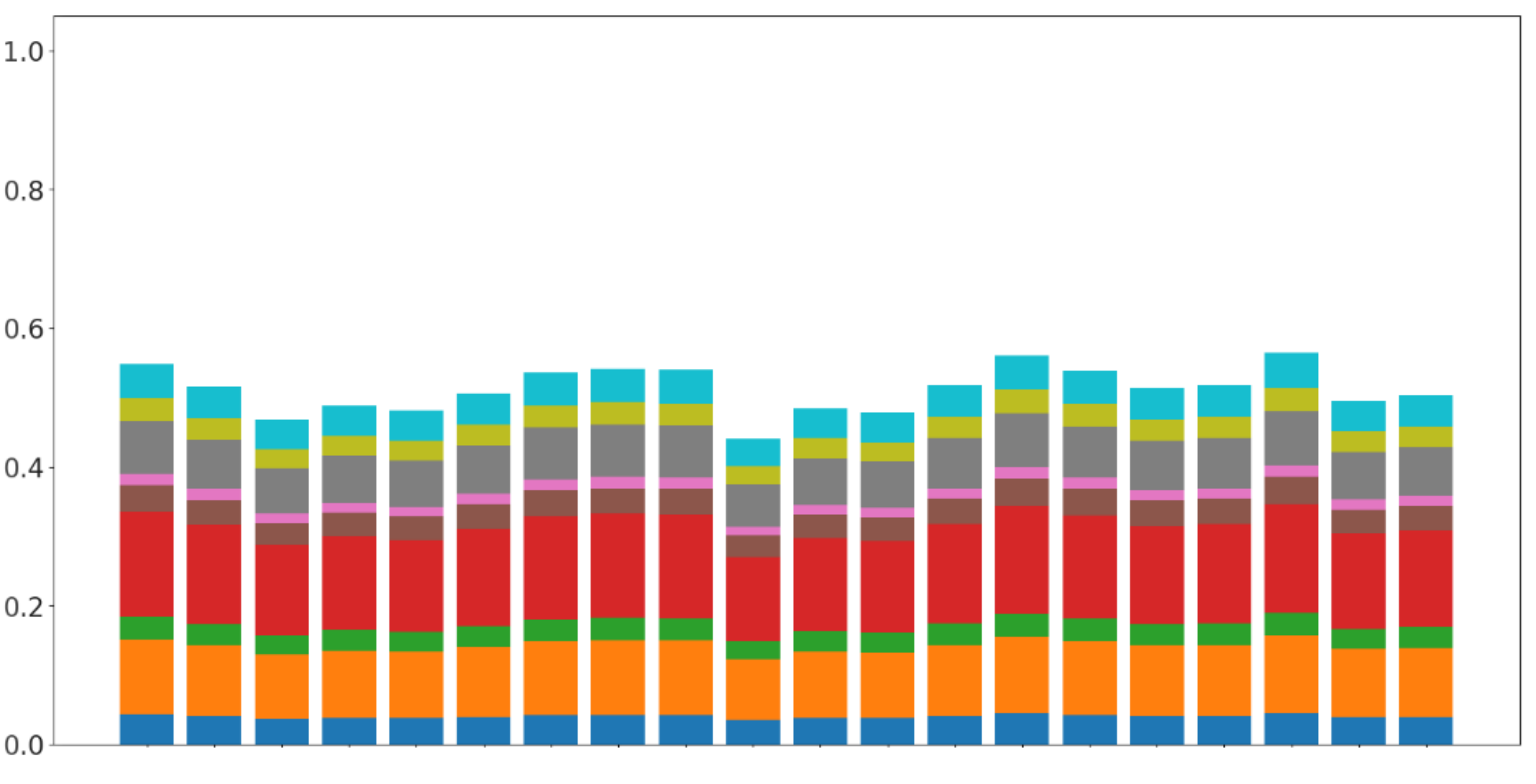}
         \caption{$\alpha=5.0, E=1$}
     \end{subfigure}
     \hfill
     \begin{subfigure}[b]{0.24\textwidth}
         \centering
         \includegraphics[width=\textwidth]{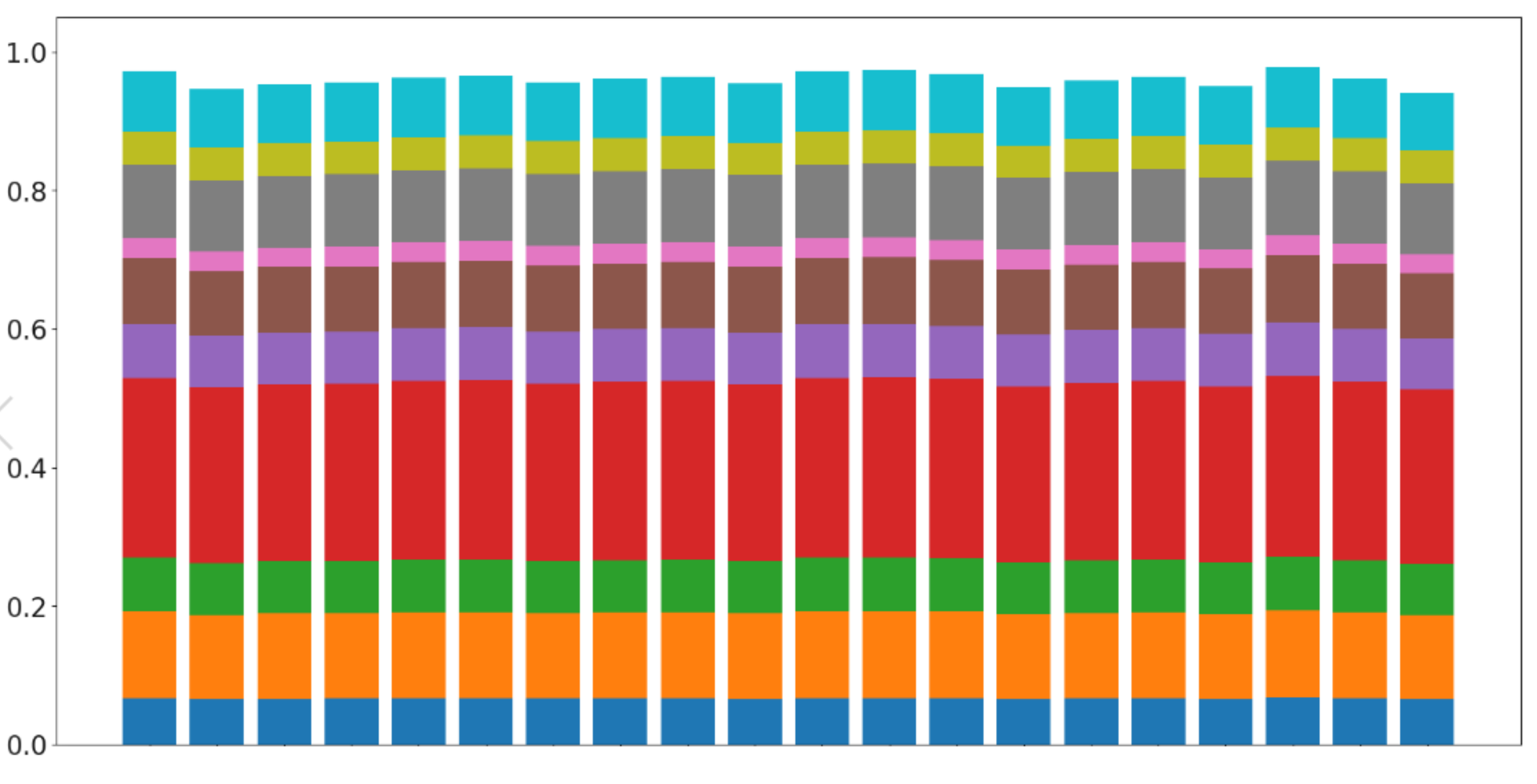}
         \caption{$\alpha=5.0, E=5$}
     \end{subfigure}
        \caption{\textbf{Top-1 per-class accuracy of FedPdp\,\citep{li2019convergence} according to $\alpha$ and $E$ on each client's data.} The x-axis indicates each client. In all results, the clients are sampled 8 out of 20. Each top and bottom row represents the evaluations with client and class heterogeneity, respectively.}
        \label{fig:perfor_fedavg_convergence}
\end{figure}

As shown in \autoref{fig:perfor_fedavg_convergence}, top1 per-class accuracy is significantly variant over all clients in the non-IID setting while the IID setting does not.
\section{Bandit Initialization}
\label{sec:bandit}
{\bf FedCM-UCB.} For the initialization of reward, $a_k$ is set to 1, and $\hat{\mu}_k$ is sampled from the random binomial distribution.

{\bf FedCM-TS.} All $\alpha_{k}, \beta_{k}$ is initialized with (1,1).

\section{Detailed Experiment Results}
\label{de-ex-re}

\paragraph{Construction of validation set.} In all experiments, we constructed the validation set by sampling 500 instances per class on balanced.

\paragraph{Hyperparameter $\mu$ of proximal term in FedProx \cite{fedprox}.} We set the weight of proximal term to 0.1.

\begin{figure}[h]
     \centering
     \begin{subfigure}[b]{0.8\textwidth}
         \centering
         \includegraphics[width=\textwidth]{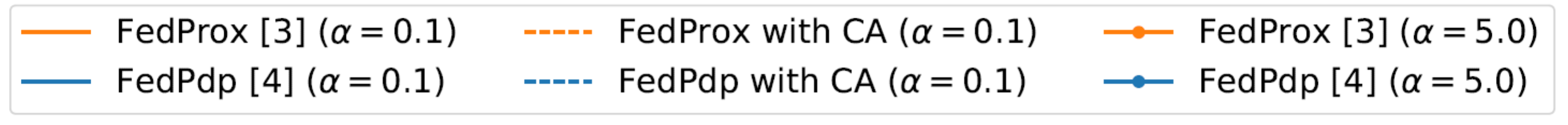}
         \vspace{-10pt}
     \end{subfigure}
     \centering
     \begin{subfigure}[b]{0.46\textwidth}
         \centering
         \includegraphics[width=\textwidth]{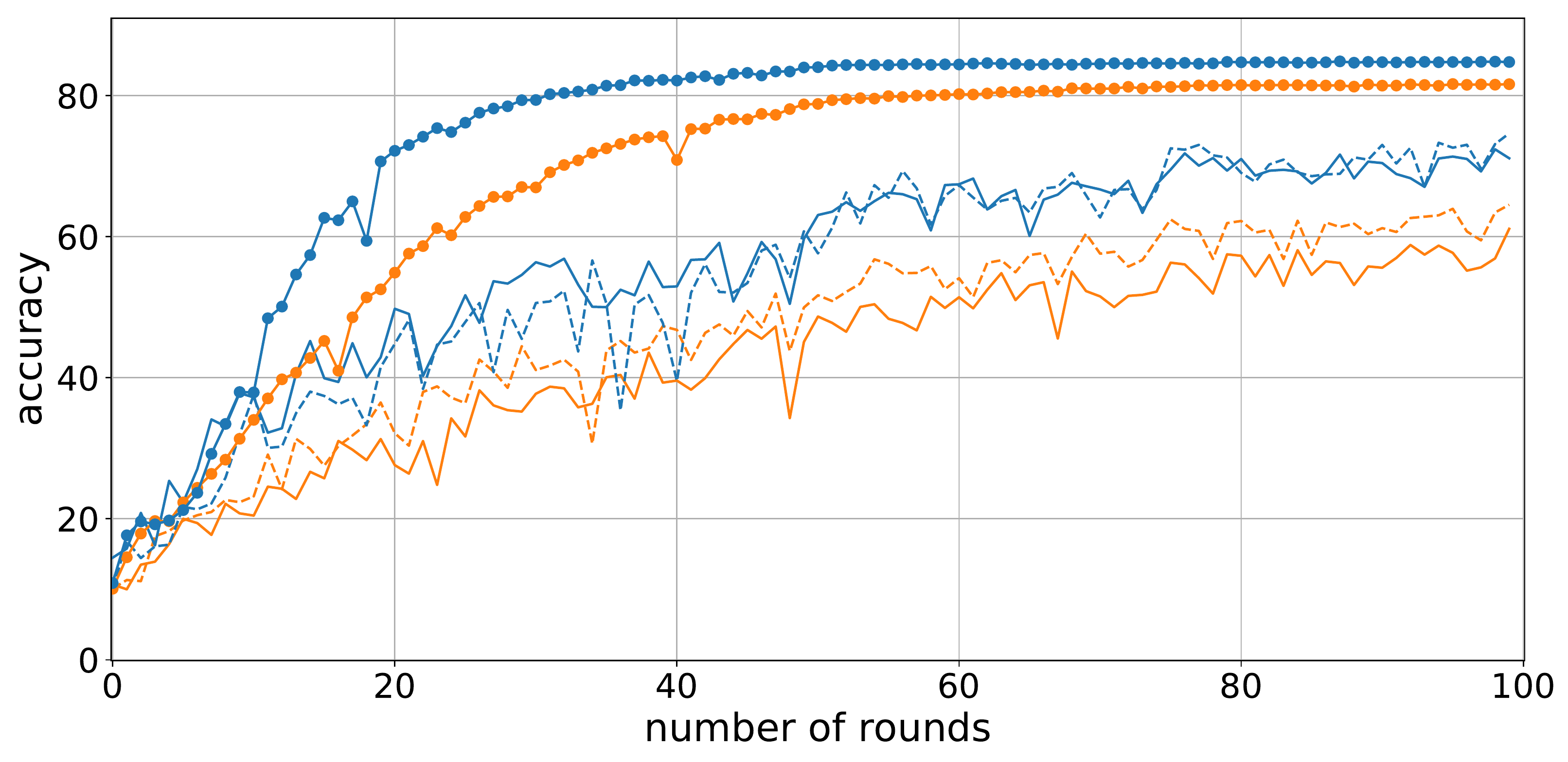}
         \caption{Client Heterogeneity}
     \end{subfigure}
     \begin{subfigure}[b]{0.46\textwidth}
         \centering
         \includegraphics[width=\textwidth]{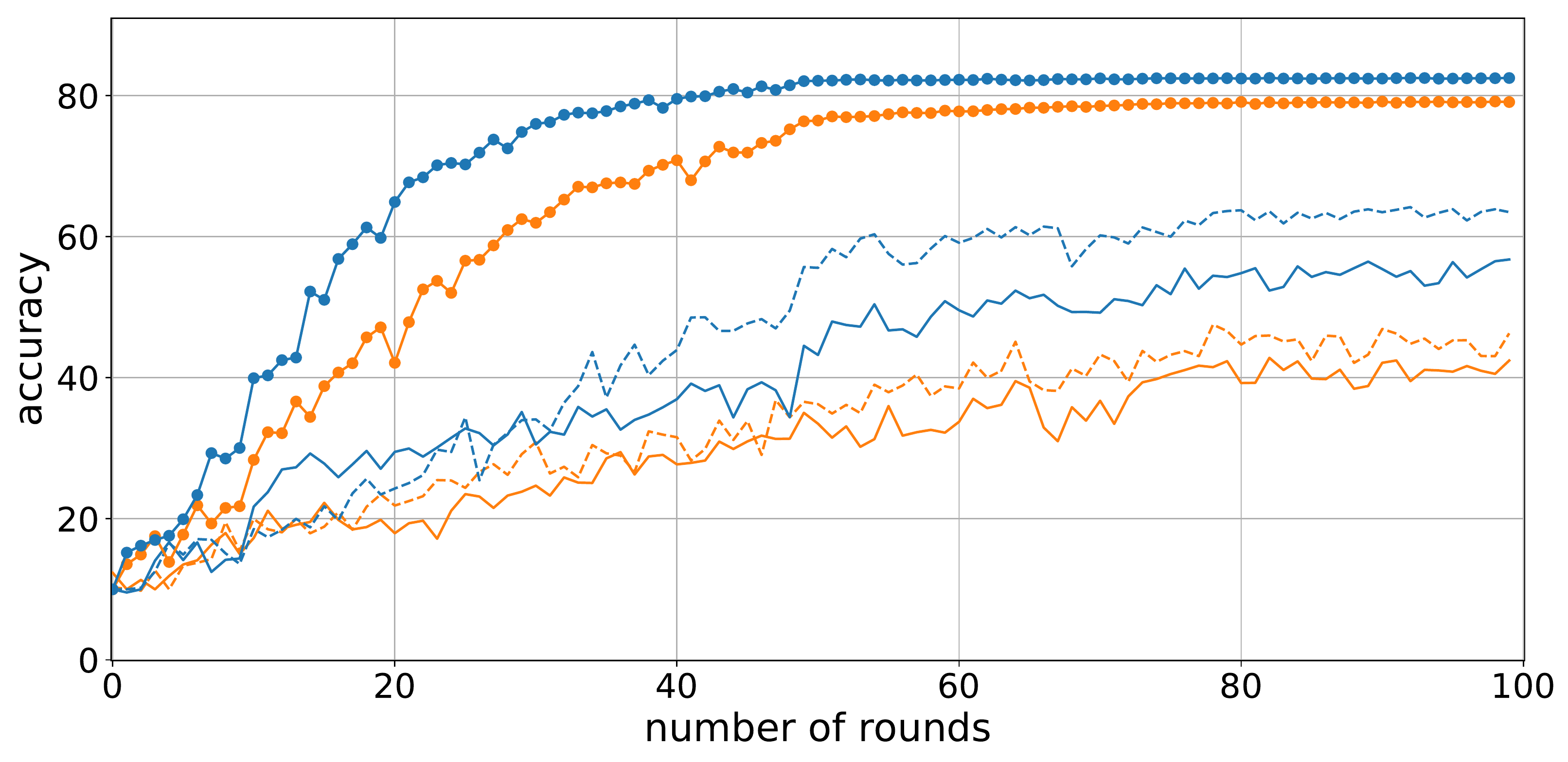}
         \caption{Class Heterogeneity}
     \end{subfigure}
        \caption{\textbf{Dataset across 20 clients according to IIDness\,($\alpha)$ and heterogeneity}. }
        \label{fig:conver_spe}
\end{figure}

\paragraph{Training accelerations according to $\alpha$.} As shown in \autoref{fig:conver_spe}, the performance of models trained with IID data was better than those with non-IID data in the conventional sampling schemes.

\paragraph{}

\paragraph{Further experimental results.} \autoref{tab:appendix_figure3detail} and \autoref{tab:appendix_figure4detail} show the further comparison of FedCA and FedCM for the conventional FL algorithms. \autoref{tab:appendix_figure5detail} shows the convergence speed of algorithms mentioned in \autoref{tab:appendix_figure4detail}.

\begin{table}[h]
\caption{\textbf{Top-1 accuracy of FedCA for FedProx \citep{fedprox} and FedPdp \citep{li2019convergence}} With sampling 8 out of 20 clients, all models are trained in $E=5$ and Non-IID\,($\alpha=0.1$) settings. The standard deviation values are calculated as results of 2 different seeds.
}
\centering
\small 
\vspace{+5pt}
\begin{tabular}{@{}ccccc@{}} 
\toprule
\multirow{2}{*}{Heterogeneity} & \multirow{2}{*}{FedCA} & \multirow{2}{*}{Score function} & \multicolumn{2}{c}{Algorithm} \\ \cmidrule(l){4-5} 
 & & & FedProx \citep{fedprox} & FedPdp \citep{li2019convergence}  \\ \cmidrule(r){1-1}\cmidrule(r){2-2}\cmidrule(r){3-3}\cmidrule(r){4-5}
\multirow{3}{*}{Client Heterogeneity} & W/O FedCA & - & 62.75 $_{\pm\textbf{1.54}}$ & 74.16 $_{\pm0.83}$ \\ \cmidrule(r){2-2}\cmidrule(r){3-3}\cmidrule(r){4-5}
& \multirow{2}{*}{W/ FedCA} & Dirac delta & 62.83 $_{\pm1.57}$ & 73.70 $_{\pm\textbf{0.26}}$ \\ \cmidrule(r){3-3} \cmidrule(r){4-5} & & Classification loss & \textbf{64.54} $_{\pm1.94}$ & \textbf{74.73} $_{\pm{2.26}}$ \\
\cmidrule(r){1-1}\cmidrule(r){2-2}\cmidrule(r){3-3}\cmidrule(r){4-5}
\multirow{3}{*}{Class Heterogeneity} & W/O FedCA & - & 44.34 $_{\pm1.95}$ & 57.75 $_{\pm\textbf{3.44}}$ \\ \cmidrule(r){2-2}\cmidrule(r){3-3}\cmidrule(r){4-5}
& \multirow{2}{*}{W/ FedCA} & Dirac delta & \textbf{50.73} $_{\pm\textbf{1.24}}$ & 60.52 $_{\pm4.73}$ \\ \cmidrule(r){3-3} \cmidrule(r){4-5} & & Classification loss & 48.33 $_{\pm2.13}$ & \textbf{62.78} $_{\pm5.96}$ \\ \bottomrule
\end{tabular}
\label{tab:appendix_figure3detail}
\end{table}

\begin{table}[ht]
\caption{\textbf{Detailed results of FedCM compared to FedProx \citep{fedprox} and FedPdp \citep{li2019convergence}} With sampling 8 out of 20 clients, all models are trained in $E=5$ and Non-IID($\alpha=0.1$) settings. The standard deviation values are calculated as results of 2 different seeds.
}
\centering
\small 
\vspace{+5pt}
\resizebox{\textwidth}{!}{
\begin{tabular}{@{}cccccccc@{}} 
\toprule
\multirow{2}{*}{Heterogeneity} & \multirow{2}{*}{FedCA} & \multirow{2}{*}{Score function} & \multicolumn{4}{c}{Algorithm} \\ \cmidrule(l){4-8} 
 & & & FedAvg \citep{fedavg} & FedProx \citep{fedprox} & FedPdp \citep{li2019convergence} & FedCM-UCB & FedCM-TS  \\ \cmidrule(r){1-1}\cmidrule(r){2-2}\cmidrule(r){3-3}\cmidrule(r){4-8}
\multirow{3}{*}{Client Heterogeneity} & W/O FedCA & & 63.97 $_{\pm{0.90}}$ & 62.75 $_{\pm1.54}$ & 74.16 $_{\pm0.83}$ & - & - \\ \cmidrule(r){2-2}\cmidrule(r){3-3}\cmidrule(r){4-8}
& \multirow{2}{*}{W/ FedCA} & Dirac delta & - & 62.83 $_{\pm1.57}$ & 73.70 $_{\pm0.26}$ & 75.05 $_{\pm{0.46}}$ & \textbf{75.06} $_{\pm{1.56}}$ \\ \cmidrule(r){3-3} \cmidrule(r){4-8} & & Classification loss & - & 64.54 $_{\pm1.94}$ & 74.73 $_{\pm{2.26}}$ & 72.05 $_{\pm{2.52}}$ & 72.95 $_{\pm{3.28}}$ \\
\cmidrule(r){1-1}\cmidrule(r){2-2}\cmidrule(r){3-3}\cmidrule(r){4-8}
\multirow{3}{*}{Class Heterogeneity} & W/O FedCA & & 40.24 $_{\pm{4.04}}$ & 44.34 $_{\pm1.95}$ & 57.75 $_{\pm3.44}$ & - & - \\ \cmidrule(r){2-2}\cmidrule(r){3-3}\cmidrule(r){4-8}
& \multirow{2}{*}{W/ FedCA} & Dirac delta & - & 50.73 $_{\pm1.24}$ & 60.52 $_{\pm4.73}$ & 63.96 $_{\pm{0.74}}$ & 61.29 $_{\pm{1.44}}$ \\ \cmidrule(r){3-3} \cmidrule(r){4-8} & & Classification loss & - & 48.33 $_{\pm2.13}$ & 62.78 $_{\pm5.96}$ & 61.76 $_{\pm{2.67}}$ & \textbf{64.13} $_{\pm{\textbf{0.6}}}$ \\ \bottomrule
\end{tabular}}
\label{tab:appendix_figure4detail}
\end{table}

\begin{table}[ht]
\caption{\textbf{Communication rounds over various alogrithms.} Here, we evaluate the convergence speed via the certain communication round that the performance reach the accuracy of FedAvg \citep{fedavg} (e.g., Client Heterogeneity: $63.97\%$, Class Heterogeneity: $40.24\%$). All training settings are the same with that of \autoref{tab:appendix_figure4detail}.}
\centering
\small 
\vspace{+5pt}
\resizebox{\textwidth}{!}{
\begin{tabular}{@{}cccccccc@{}} 
\toprule
\multirow{2}{*}{Heterogeneity} & \multirow{2}{*}{FedCA} & \multirow{2}{*}{Score function} & \multicolumn{4}{c}{Algorithm} \\ \cmidrule(l){4-8} 
 & & & FedAvg \citep{fedavg} & FedProx \citep{fedprox} & FedPdp \citep{li2019convergence} & FedCM-UCB & FedCM-TS  \\ \cmidrule(r){1-1}\cmidrule(r){2-2}\cmidrule(r){3-3}\cmidrule(r){4-8}
\multirow{3}{*}{Client Heterogeneity} & W/O FedCA & & 100 (1$\times$) & - & 52 (1.92$\times$) & - & - \\ \cmidrule(r){2-2}\cmidrule(r){3-3}\cmidrule(r){4-8}
& \multirow{2}{*}{W/ FedCA} & Dirac delta & - & - & 57 (1.75$\times$) & 35 (2.86$\times$) & \textbf{31 (3.23$\times$)} \\ \cmidrule(r){3-3} \cmidrule(r){4-8} & & Classification loss & - & 84 (1.19$\times$) & 49 (2.04$\times$) & 53 (1.87$\times$) & 51 (1.96$\times$) \\
\cmidrule(r){1-1}\cmidrule(r){2-2}\cmidrule(r){3-3}\cmidrule(r){4-8}
\multirow{3}{*}{Class Heterogeneity} & W/O FedCA & & 100 (1$\times$) & 71 (1.41$\times$) & 46 (2.17$\times$) & - & - \\ \cmidrule(r){2-2}\cmidrule(r){3-3}\cmidrule(r){4-8}
& \multirow{2}{*}{W/ FedCA} & Dirac delta & - & 52 (1.92$\times$) & 35 (2.86$\times$) & 27 (3.70$\times$) & \textbf{24 (4.17$\times$)} \\ \cmidrule(r){3-3} \cmidrule(r){4-8} & & Classification loss & - & 59 (1.69$\times$) & 34 (2.94$\times$) & 33 (3.03$\times$) & 25 (4.00$\times$) \\ \bottomrule
\end{tabular}}
\label{tab:appendix_figure5detail}
\end{table}

\end{CJK}
\end{document}